\documentclass{article} 
\usepackage{arxiv}
\usepackage{natbib} 

\usepackage{amsmath,amsfonts,bm}

\def\eqref#1{equation~\ref{#1}}

\def\1{\bm{1}}

\DeclareMathAlphabet{\mathsfit}{\encodingdefault}{\sfdefault}{m}{sl}
\SetMathAlphabet{\mathsfit}{bold}{\encodingdefault}{\sfdefault}{bx}{n}

\newcommand{\R}{\mathbb{R}}

\usepackage{mathtools}
\usepackage{hyperref}
\usepackage{url}
\usepackage{multirow}
\usepackage{graphicx}
\usepackage{amsmath}
\usepackage{derivative}
\usepackage[T1]{fontenc}
\title{An Analytical Theory of Auxiliary Learning}

\author{Federico Milanesio \\
Department of Physics and INFN\\
University of Turin\\
Via Giuria 1, 10125 Turin, Italy\\
\texttt{federico.milanesi@unito.it} \\
\And
Alessandro Ingrosso \\
Donders Centre for Neuroscience \\
Radboud University\\
Heyendaalseweg 135, 6525 AJ Nijmegen, The Netherlands\\
\texttt{alessandro.ingrosso@donders.ru.nl} \\
\And
Matteo Osella \\
Department of Physics and INFN\\
University of Turin\\
Via Giuria 1, 10125 Turin, Italy\\
\texttt{matteo.osella@unito.it} \\
}

\usepackage{xstring}
\usepackage{etoolbox}

\newcommand{\trimurl}[1]{%
  \IfBeginWith{#1}{https://}%
    {\StrGobbleLeft{#1}{8}[\tempurl]}%
    {\IfBeginWith{#1}{http://}%
      {\StrGobbleLeft{#1}{7}[\tempurl]}%
      {\def\tempurl{#1}}}%
  \href{#1}{\nolinkurl{\tempurl}}}

\AtBeginEnvironment{thebibliography}{\let\url\trimurl}

\begin{document}

\maketitle
\begin{abstract}

   Auxiliary learning is an optimization paradigm in which a neural network's performance on a target task is improved by jointly training it on additional tasks. However, the mechanisms behind this improvement remain poorly understood. We study this problem using a teacher-student framework and derive a closed system of differential equations describing the dynamics of online stochastic gradient descent in the large-input limit. For linear networks, we obtain a closed-form expression for the generalization error to leading order in the learning rate, quantifying how task correlations and label noise determine the benefit of auxiliary learning. For non-linear activation functions, we develop a fluctuation-dissipation analytical theory that establishes a general relation linking the main and auxiliary errors to the corresponding single-task error. Numerical experiments support the theoretical predictions and show how auxiliary tasks improve generalization by balancing the forcing dynamics towards the optimal solution with gradient noise.
   
    \end{abstract}

\section{Introduction}

Modern neural networks are rarely trained to optimize a single task in isolation. In practice, practitioners have introduced different multi-task paradigms to improve performance on the tasks these models are designed for \citep{caruana_1997,ruder_2017, liebel_2018, crawshaw_2020, zhang_2021}. These paradigms use additional objectives, called \textit{auxiliary} tasks, on which the model is trained either in parallel with its main objective (auxiliary learning) or by alternating tasks (knowledge transfer and continual learning). The empirical effectiveness of these methods clashes with the intuition that multiple competing objectives might worsen the overall performance. 

Modern neural networks are prone to overfitting, but do not have a capacity problem, since they can approximate any function to any precision. \citet{baxter_2000} suggested that multi-task learning can work as an \textit{inductive bias} in the space of possible functions. Thus, multi-task learning might lead the network to learn a better-generalizing representation of the data, but our understanding of this phenomenon is still limited.

To begin developing a theory of auxiliary learning,  we focus on a specific tractable setting. We study a variant of the classic teacher-student setup in which one or more auxiliary tasks, available only during training, can improve performance on a target task. In this framework, the target function is exactly learnable, and we analyze how auxiliary tasks affect generalization through training dynamics, while leaving the target-task optimal solution unchanged. To do so, we expand on the framework proposed for the single-output teacher-student setup by \citet{saad_1995a, saad_1995b, biehl_1995}, who showed that a set of ordinary differential equations can describe the dynamics of large-width networks trained with online learning. We generalize these mean-field ODEs to our auxiliary problem and, drawing upon the literature on adaptive filters, we derive a closed-form leading-order approximation of the generalization error for a multi-layer linear model.

We then focus on other nonlinear activations and find that, at convergence, the dynamics are described, to first order, by a Lyapunov equation. By solving it numerically, we can recover the generalization error, as well as the optimal weighting of the auxiliary task that minimizes it, for any choice of activation and hyperparameters. We find that the generalization error between the main and auxiliary task is described by a simple relationship involving the generalization error of the corresponding one-output teacher-student network, which our theory predicts for any choice of activation function; a novel result, since the existing literature recovers it only for some specific activations \citep{goldt_2020}.

\subsection{Previous literature}

\paragraph{Auxiliary learning.} Auxiliary learning uses parallel tasks during training as an inductive bias to improve either feature learning, generalization, or optimization. \citet{caruana_1997} introduced the idea that related tasks improve generalization via learning shared representations. \citet{baxter_2000} later expanded on this relationship. Its effectiveness has been extensively discussed; \citet{caruana_1997,ruder_2017, liebel_2018, crawshaw_2020, zhang_2021} provide relevant literature reviews

\paragraph{Adaptive filters.} Adaptive filters are linear functions, explicitly designed as real-time, sample-by-sample algorithms for processing streaming signals, where parameters are updated after each input signal is processed. In the adaptive filters literature, the steady-state error of the least mean squares algorithm (which is exactly online stochastic gradient descent, SGD, on linear regression with quadratic loss) has been studied extensively by \citet{widrow_1983, haykin_1986, sayed_2008}.

\paragraph{Teacher-student.} A series of works on the teacher-student setup \citep{saad_1995a, saad_1995b, biehl_1995, vicente_1997, saad_1999, inoue_2003} showed that in the large-input limit, when the size of the input data is much larger than the number of hidden neurons ($N\to\infty$), the dynamics of the student network trained with online stochastic gradient descent (SGD) can be modeled by a set of mean-field ordinary differential equations (ODEs) for some order parameters, whose number scale intensively as $K^2$, where $K$ is the number of neurons in the hidden layer, and do not depend on the (extensive) size of the input $N$. This statistical mechanics framework, one of the first used to study the properties of multi-layer neural networks, was later formalized by \citet{goldt_2020} and adopted to understand specific settings related to multi-task learning; \citet{straat_2019} expanded it to ReLU activations, \citet{asanuma_2021, goldt_2021, saglietti_2022} to continual learning, and \citet{mori_2025} to dropout regularization. \citet{citton_2024, citton_2025, citton_2025b} have used Hermite expansions to simulate the dynamics of arbitrary activation functions.

\paragraph{Fluctuation-dissipation.} \citet{mandt_2017} expanded the model loss around a local minimum, so that SGD reduces to a multivariate Ornstein-Uhlenbeck process whose Gaussian stationary distribution has a covariance matrix fixed by a Lyapunov equation involving the Hessian at the optimum. Several works have expanded this first result; for example, \citet{jastrzebski_2018} solved the stationary covariance in closed form under additional assumptions. \citet{yaida_2018} gives a more general discussion of fluctuation-dissipation relations without the quadratic approximation.

\section{Auxiliary learning in the teacher-student framework}

\subsection{Problem setup}

In the teacher-student setup, a \textit{student} network is trained on inputs $\boldsymbol x \in \mathbb{R}^N$ drawn from the normal distribution $\mathcal{N}(0, I_N)$ with labels generated by a \textit{teacher} network. We consider a variant of the classic teacher-student scenario in which a task (here called the main task) must be learned, with the help of an auxiliary output. The setting is thus that of a multi-output network with $K$ hidden neurons,  parametrized by weights $W\in\mathbb{R}^{K\times N},\,\boldsymbol v\in\mathbb{R}^{K},\,\boldsymbol u\in\mathbb{R}^{K}$ and activation function $g:\mathbb{R}\to\mathbb{R}$. The network outputs are

\begin{equation}
\begin{aligned}
    y_{\text{main}}(\boldsymbol{x}\,|\,\theta)= \sum_{j=1}^K v_{j}\,g\left(h_j\right)= \sum_{j=1}^K v_{j}\,g\left(\dfrac{\boldsymbol w_j\cdot\boldsymbol x}{\sqrt N}\right),\\
    y_{\text{aux}}(\boldsymbol{x}\,|\,\theta)= \sum_{j=1}^K u_{j}\,g\left(h_j\right)= \sum_{j=1}^K u_{j}\,g\left(\dfrac{\boldsymbol w_j\cdot\boldsymbol x}{\sqrt N}\right),\\
\end{aligned}
\end{equation}

where $\boldsymbol w_j$ is the j-th row of $W$. We call $\theta$ the set of weights $\{W,\boldsymbol v,\boldsymbol u\}$. The teacher and student networks have the same number of neurons. The main target is $y_{\text{main}}$, and $y_{\text{aux}}$ is the auxiliary target.

Given a set of teacher weights $\theta^*=\{W^*,\boldsymbol v^*,\boldsymbol u^*\}$, the labels to be learned are $ y^*(\boldsymbol{x}) = y (\boldsymbol{x}\,|\,\theta^*)$, and we are interested in the generalization error on the target task

\begin{equation}
    \epsilon_{\text{main}} (\theta, \theta^*)= \underset{\boldsymbol x\in \mathcal{N}}{\mathbb{E}}\bigg[\bigg( y^*_{\text{main}} (\boldsymbol{x})-y_{\text{main}} (\boldsymbol{x})\bigg)^2\bigg].
\end{equation}

\begin{figure*}
\centering
\includegraphics[width=0.85\linewidth]{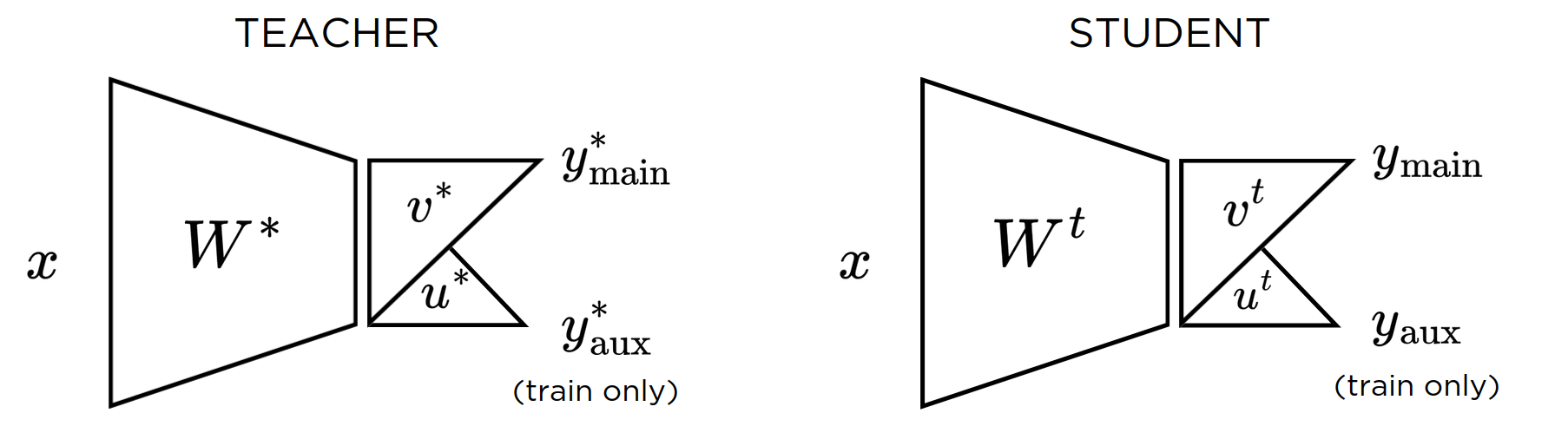}
 \caption{\textbf{Schema of the multi-index teacher-student setup.}}
\end{figure*}

 We initialize the student model weights from a standard Gaussian (but they can be initialized differently without loss of generality). The student is trained with online SGD: namely, at each epoch, we sample a point from the standard normal $\boldsymbol x \sim \mathcal{N}(0, I_N)$, and the weights are updated by

\begin{equation}
\label{eq:sgd}
\begin{aligned}
    W^{t+1}=W^t-\mu\nabla_W \mathcal{L}(\boldsymbol x\,|\,\theta^t,\theta^*),\\
    v^{t+1}=v^t-\frac\mu N\nabla_v \mathcal{L}(\boldsymbol x\,|\,\theta^t,\theta^*),\\
    u^{t+1}=u^t-\frac\mu N\nabla_u \mathcal{L}(\boldsymbol x\,|\,\theta^t,\theta^*),
\end{aligned}
\end{equation}

with learning rate $\mu>0$ and training loss    

\begin{equation}
\begin{aligned}
    \Aboxed{\mathcal{L}(\boldsymbol x)&=\alpha\mathcal{L}_{\text{main}}(\boldsymbol x)+\beta\mathcal{L}_{\text{aux}}(\boldsymbol x),}\\\mathcal{L}_i(\boldsymbol x)&=\frac{1}{2}\left( y_i ^*(\boldsymbol{x})-y_i (\boldsymbol{x})+\sigma z_i\right)^2,\quad i\in\{\text{main},\text{aux}\}
\end{aligned}
\end{equation}

with $\sigma\in\mathbb{R}$, and $z_i \sim \mathcal{N}(0, 1)$ representing Gaussian noise added to the labels during training. Notice how the auxiliary loss does not change the minimizer, since the two tasks share the same teacher. As we will see, it plays a crucial dynamical role in trading effective label noise against effective learning rate.

A central question is how much to weight the auxiliary loss term to minimize the generalization error on the target task. In other words, we want to find the optimal values for the hyperparameters 

\begin{align}
(\alpha,\beta)=\underset{\alpha,\beta\in S}{\text{argmin}}\, \epsilon_{\text{main}},
\end{align}

where $S$ is some set of conditions on $(\alpha,\beta)$. We focus mainly on finding the optimal $\beta$ with $\alpha=1$, without rescaling the main generalization loss. We also explore the case  $\alpha=1-\beta$ in Appendix \ref{app:second-case}, which produces different results because of the dynamical effects governing our online learning setup.

\subsection{The large-input limit}

We train the student network on $i.i.d.$ Gaussian samples drawn from a normal distribution. Thus, both the learning dynamics and the generalization error depend on the student and teacher weights only through the statistics of the network pre-activations (local fields) $h_i$ and $h_n^*$. Since each pre-activation is a projection of the Gaussian sample onto the corresponding weight $\boldsymbol w$, the pre-activations are jointly Gaussian with zero mean, and their covariance is fully determined by

\begin{align}
Q_{ik} &= \mathbb{E}\left[h_ih_k\right]=\dfrac{\boldsymbol w_i^T \boldsymbol w_k}{N},\nonumber\\
R_{in} &=\mathbb{E}\left[h_ih_n^*\right]= \dfrac{\boldsymbol w_i^T \boldsymbol w_n^*}{N},\\
T_{mn} &=\mathbb{E}\left[h^*_mh^*_n\right]= \dfrac{\boldsymbol w_m^{*T} \boldsymbol w_n^*}{N}.\nonumber
\end{align}

The entries of these matrices, the student-student overlaps $Q_{ik}$, the student-teacher overlaps $R_{in}$, and the teacher-teacher overlaps $T_{mn}$, are the (slightly improperly named) \textit{order parameters} of our problem, and are the only quantities through which the weights enter both the generalization error and the dynamics. 

One can then compute the generalization error analytically \citep{saad_1995a, saad_1995b, biehl_1995, goldt_2020}, obtaining 

\begin{equation}
\begin{aligned}
    \epsilon_{\text{main}} = \sum_{i,j}v_iv_jI_2(i,j)+\sum_{n,m}v_n^*v_m^*I_2(n,m)-2\sum_{i,n}v_iv_n^*I_2(i,n),
\end{aligned}
\end{equation}

where the notation $I_2(i,j)$ indicates $\mathbb{E}[g(h_i)\,g(h_j)]$, with $i$, $j$ being indices of units in the student network and $n$, $m$ teacher ones. Because $I_2$ only depends on the order parameters, it reduces to computing a two-dimensional Gaussian average of a nonlinear function, which, depending on the activation, can be solved analytically or approximated numerically.

The order parameters, together with the second-layer weights $\boldsymbol v$ and $\boldsymbol u$, can be evolved through the exact population dynamics of SGD (namely, in the $N \to \infty$ limit) according to a closed set of ordinary differential equations that depend on $I_2$ and other integrals of the activation function. This description becomes deterministic and asymptotically exact in the large-$N$ limit, since the order parameters are self-averaging \citep{goldt_2020}. We derive the full set of equations for the two-output case in Appendix \ref{app:diff_eq}.

\section{Results}
\subsection{Linear model}

Before studying the general case of networks with nonlinear activation functions, we develop a theory for the simpler deep linear model, defined by the choice $g(x)=x$, with fixed second-layer weights $\boldsymbol v=\boldsymbol v^*$, and $\boldsymbol u=\boldsymbol u^*$, leaving $W$ as the only trainable parameters. Despite its simplicity, this model still provides insight into the dynamics of one-hidden-layer models. It is well understood that a linear model trained with online SGD on an MSE loss converges to a generalization error governed by the dynamical effects of training \citep{widrow_1983, haykin_1986, sayed_2008}. These linear models were introduced in the machine learning literature as \textit{adaptive filters}. For normal Gaussian data, the loss converges to $\epsilon_{a.f.} \approx \mu\, \sigma^2 / 2$ when we keep only the leading term in $\mu$. When either $\mu$ or $\sigma$ goes to zero, we recover the original function. The dependence on $\mu$ explicitly highlights the dynamical nature of this result.

\begin{figure*}[!ht]
    \centering
    \includegraphics[width=\linewidth]{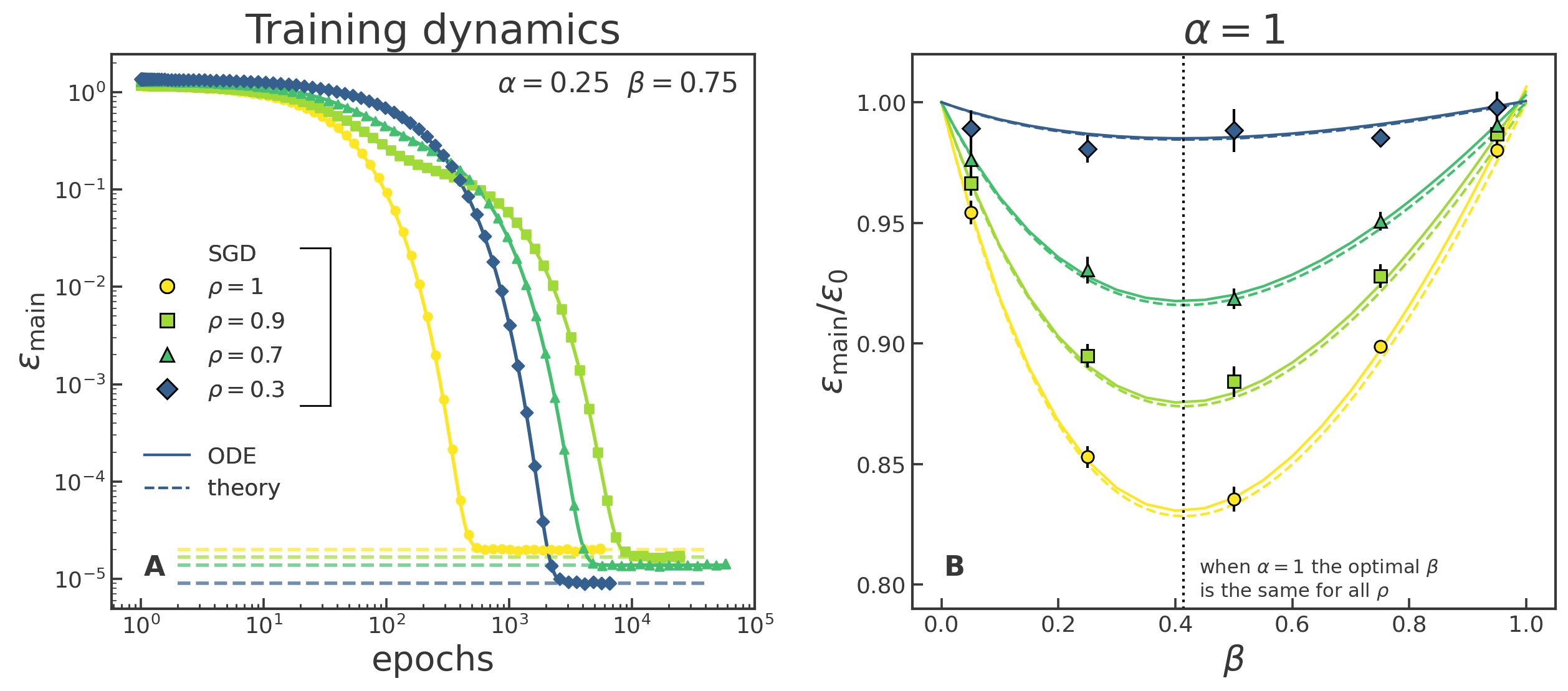}
    \caption{\textbf{Performance of linear models.} A) Sample learning trajectory obtained with ODEs (continuous line) and averages over ten runs of SGD (dots) when $\alpha=0.25$ and $\beta=0.75$. The dashed line is the theoretical loss at convergence, and different colors/shapes correspond to different correlations $\rho$. B) Relative error $\epsilon_{\text{main}}/\epsilon_0$ for a linear model with four different $\rho$ trained with online gradient descent, $\alpha=1$, and different values of $\beta$. The dashed lines show the theoretical solutions, the continuous lines show the results of numerically solving the mean-field ODEs, the dots show simulation results for models trained with online SGD, averaged over 10000 epochs after convergence across 10 simulations, with the vertical bars denoting the corresponding confidence intervals. Parameters for all simulations are $\mu=0.01$, $\sigma=0.1$, $K=4$, $N=784$.}
    \label{fig:1}
\end{figure*}

In Appendix \ref{app:linear}, we show that, for $\alpha>0$, $\beta\geq0$,  a small learning rate $\mu$, and with the same norms of the second layer weights (namely,  $|\boldsymbol u|^2 = |\boldsymbol v|^2$), the two-layer linear model converges to a generalization loss with a leading term in $\mu$

\begin{equation}
\begin{aligned}
\label{err_linear}
    \epsilon_{\text{main}} &= \dfrac{\mu \sigma^2 }{2}|\boldsymbol v|^2\dfrac{\alpha^2+\alpha\beta(1-\rho^2)+\rho^2\beta^2}{\alpha+\beta}=\\
    &=\underbrace{\epsilon_{a.f.} |\boldsymbol v|^2}_{\epsilon_0} \dfrac{\alpha^2+\alpha\beta(1-\rho^2)+\rho^2\beta^2}{\alpha+\beta},
\end{aligned}
\end{equation}

where $\rho ={\boldsymbol v\cdot\boldsymbol u}/{|\boldsymbol v||\boldsymbol u|}$ is the correlation between the two output weights and $\epsilon_{a.f.}$ is the generalization error of the corresponding linear model trained without an auxiliary task (also truncated at the first order in $\mu$). In Figure \ref{fig:1}A, we plot the evolution of generalization error $\epsilon_{\text{main}}$ obtained both by training models with SGD and by numerically solving the ODEs, showing how their dynamics match and converge to the loss predicted by the theory for small learning rates $\mu$. For SGD, the generalization error fluctuates around the steady state, so we show an average over 10000 epochs after convergence.

When setting $\alpha=1$, we get

\begin{equation}
\label{eq:fixed_alpha_lin}
    \epsilon_{\text{main}}\Big|_{\alpha=1}=\epsilon_0\left(1-\rho^2\dfrac{\beta(1-\beta)}{1+\beta}\right),
\end{equation}

which is minimized by

\begin{equation}
    \beta^*=\sqrt2-1.
\end{equation}

Interestingly, the optimal value does not depend on $\rho$, as shown in Figure \ref{fig:1}B. When $\rho=0$, the generalization error is $\epsilon_{\text{main}}=\epsilon_{a.f.}|\boldsymbol{v}|^2$ and does not depend on the choice of $\beta$: when the main and auxiliary tasks are completely decoupled, rescaling $\beta$ does not rescale the main task learning rate and does not affect its generalization error. When the two tasks coincide (namely, when $\rho=1$), the optimal value  $\beta^*$ is not $1/2$ as one naively expects. As shown in Appendix \ref{app:comm}, the case of $\rho=1$ corresponds to  an effective learning rate $\mu_\text{eff}=\mu(\alpha+\beta)$ and an effective noise $\sigma^2_\text{eff}=\sigma^2{(\alpha^2+\beta^2)}/{(\alpha+\beta)^2}$. Thus, rescaling the auxiliary task also increases the effective learning rate, creating a trade-off between the learning rate and the noise variance. 

A different phenomenology can be obtained with the constraint $\alpha=1-\beta$, when $\mu_\text{eff}=\mu$, without a dependence on $\beta$. We discuss this setting in Appendix \ref{app:second-case}, and show how this additional constraint modifies the results.

Quite unexpectedly, from Equation \ref{err_linear} it follows that

\begin{equation}
\label{eq:loss_relationship}
    \alpha\epsilon_{\text{main}}+\beta\epsilon_{\text{aux}}= \epsilon_{a.f.}|\boldsymbol v|^2(\alpha^2+\beta^2),
\end{equation}

where the quantities on the left depend on $\rho$, while those on the right do not; we return to this relationship when discussing nonlinear models.

In Appendix \ref{app:linear}, we extend these results to a teacher with different noise variance on the auxiliary task, showing it is beneficial to weight the auxiliary task more when its noise variance is smaller than the main task one, as well as a teacher with multiple auxiliary tasks, showing that they improve generalization compared to only having one.

\subsection{Nonlinear activation functions}\label{sec:nonlinear}

\citet{goldt_2020} previously calculated the generalization error for Soft Committee Machines (SCMs), networks for which $v_i=v^*_i=1\;\forall i$, showing it to be $\epsilon_\text{erf}=\mu \sigma^2K /(\pi\sqrt 3)+\mathcal{O}(\mu^2)$ when the activation function is  $g(x)=\text{erf}(x/\sqrt2)$. This function is convenient because it is one of the few for which all ODE integrals are analytically computable in a simple form. In general, they also observe that $\epsilon_0\propto \mu \sigma^2$ when $\mu\to0$ with ReLU activation, although they did not derive the proportionality constant analytically. 

\begin{figure*}[!ht]
    \centering
    \includegraphics[width=\linewidth]{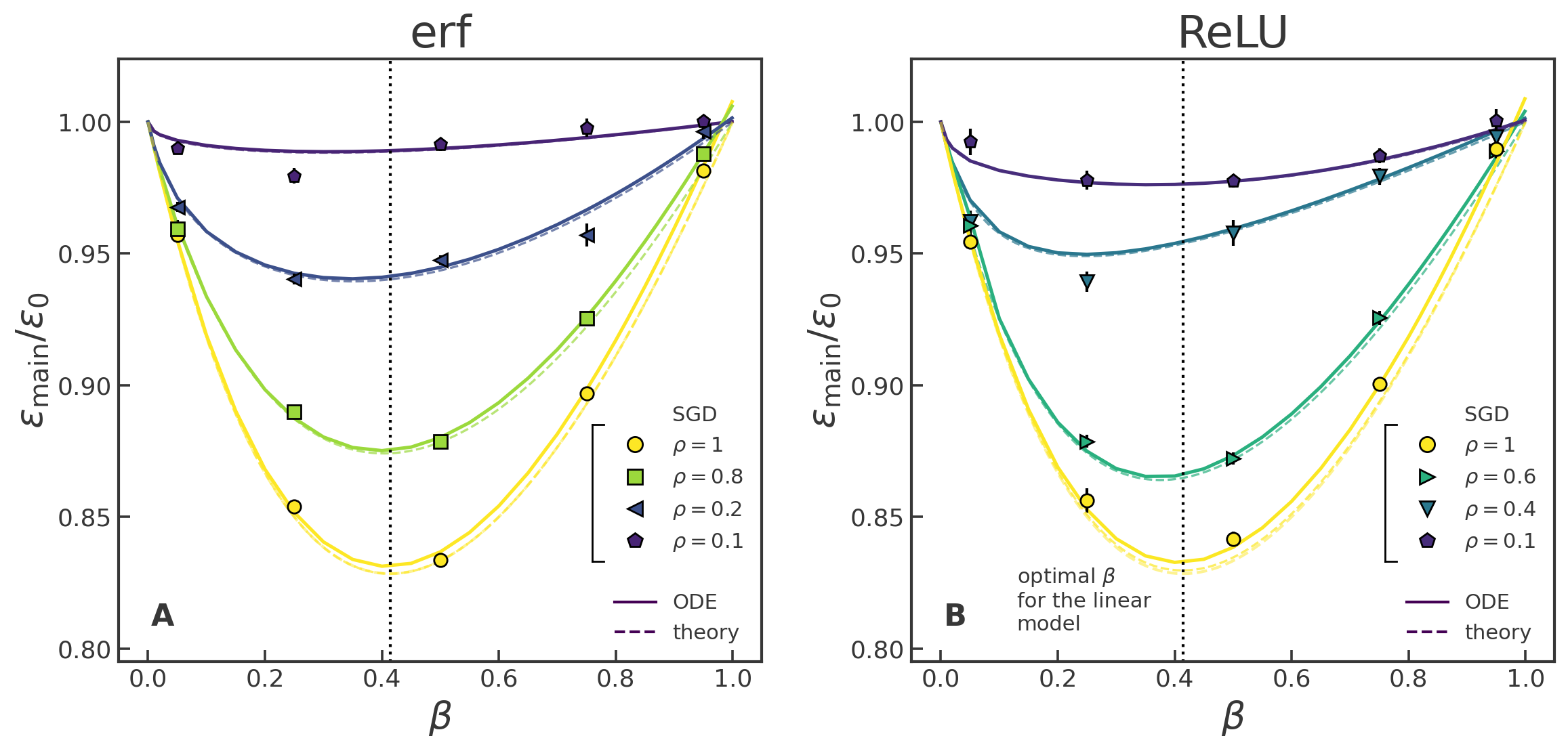}
    \caption{\textbf{Performance of models with different activations.} Relative error $\epsilon_{\text{main}}/\epsilon_0$ for a model with A) erf activation ($K=4$) and B) ReLU activation ($K=2$), with different $\rho$ trained with online gradient descent, $\alpha=1$, and different $\beta$s. The dashed lines show the theoretical solutions, the continuous lines show the results of numerically solving the mean-field ODEs, the dots show simulation results for models trained with online SGD, averaged over 10000 epochs after convergence across 5 simulations for erf and 10 for ReLU, with the vertical bars denoting the corresponding confidence intervals. Results for $\mu=0.01$, $\sigma=0.1$, $N=784$.}
    \label{fig:2}
\end{figure*}

We now extend these results to networks trained with an auxiliary output. We fix $\boldsymbol v=\boldsymbol v^*$ and $\boldsymbol u=\boldsymbol u^*$, with correlation $\rho$, and same norm $|\boldsymbol u|=|\boldsymbol v|$, and different nonlinear activations, namely erf and ReLU. The behavior is more complicated and does not depend only on $\rho$. Figure \ref{fig:2}A shows simulation results for erf, and Figure \ref{fig:2}B for ReLU.

\textit{A note on initialization.} Throughout the rest of this paper, we are interested in studying the behavior of the well-converging initializations $W_0$ that reach a neighbourhood of the realizable optimum $W^*$ without getting stuck in non-optimal minima or saddle points. The theoretical construction of the mean-field ODEs does not depend on the choice of initialization for $W$, $v$, and $u$, justifying the choice of specific initializations. We empirically observe that bad initializations converge to a non-diagonal $R$ matrix (possibly transiently). We also see that some of these bad initializations do not converge for $\beta=0$, namely the single-output case, but do reach a neighbourhood of the optimum in the presence of auxiliary tasks, when $\beta\neq0$. This result suggests that auxiliary learning can also help optimization, beyond the convergence results which are the focus of our theory. Appendix \ref{app:convergence} presents a preliminary computational analysis of this effect, but a complete study of how auxiliary learning might help escape bad minima in the teacher-student is beyond the scope of this work.

We can make a few empirical observations from Figure \ref{fig:2}.

\paragraph{Observation 1.} \textit{The generalization error  is $\epsilon_{\text{main}}|_{\alpha=\beta=1}=\epsilon_0$, exactly as in the linear case. }

\paragraph{Observation 2.} \textit{The relation $\alpha\epsilon_{\text{main}}+\beta\epsilon_{\text{aux}}\approx \epsilon_0(\alpha^2+\beta^2)$ still approximately holds. While both $\epsilon_{\text{main}}$ and $\epsilon_{\text{aux}}$ depend on the specific choice of $\boldsymbol v$, $\boldsymbol u$ and their correlation $\rho$, the quantity on the right side of the equation does not.}

\paragraph{Observation 3.} \textit{Empirically, the optimal value of $\beta$ is  $\beta^*<1/2$  when $\alpha=1$, and typically $\beta^* \leq \sqrt 2 - 1$ for most sampled configurations.}

\subsection{A general analytic theory}

SGD dynamics, after converging, fluctuate around the steady state given by the realizable optimum $W^*$, namely the teacher weights.  We now discuss how the statistics of this stationary distribution around the minimum allow us to compute the generalization error and explain the observations in the previous section. In Appendix \ref{app:lyap}, we show how the distribution induced by online SGD is described at first order in $\mu$ by the Lyapunov equation

\begin{equation}HC+CH=\mu\,\Sigma_{\text{grad}},\label{eq:lyap}\end{equation}

with

\begin{equation}
\begin{gathered}  C=\mathbb E \big[\delta  \boldsymbol W  \delta\boldsymbol W^T\big],\quad H=\mathbb E\big[\nabla^2_{\boldsymbol W}{\mathcal L}\big]\big|_{\boldsymbol W^*},\\
\Sigma_{\text{grad}}=\mathbb E\big[\nabla_{\boldsymbol W}{\mathcal L}\,\nabla_{\boldsymbol W}{\mathcal L}^T\big]\big|_{\boldsymbol W^*},
\end{gathered}
\end{equation}

where the weight fluctuations are $\delta \boldsymbol W:=\boldsymbol W-\boldsymbol W^*$, and we adopted the notation $\boldsymbol W = \text{vec} (W)$. Also, $\mathbb E$ is the population average over new samples once the dynamics converge around the minimum. Notice that $H$ and $\Sigma_{\text{grad}}$ are fixed by the choice of network architecture and teacher weights. 

One can numerically solve Equation \ref{eq:lyap} for $C$ and recover the theoretical generalization error, but $C$ is still a $KN\times KN$ matrix. In Appendix \ref{app:lyap2}, we show that explicitly computing $H$ and $\Sigma_{\text{grad}}$, due to the tensor product structure, reduces the Lyapunov equation to a $K\times K$ equation

\begin{equation}\label{eq:lyap_P}(\alpha P_\text{main}+\beta P_\text{aux})\tilde C+\tilde C(\alpha P_\text{main}+\beta P_\text{aux})=\mu\sigma^2(\alpha^2P_\text{main}+\beta^2P_\text{aux}).\end{equation}

with

\begin{equation}
\begin{aligned} (P_\text{main})_{ij}&=v_iv_j\,\mathbb E\big[g'(h_i)\,g'(h_j)\Big]\Big|_{W^*}=v_iv_j\,J_2(ij)\big|_{W^*},\\(P_\text{aux})_{ij}&=u_iu_j\,\mathbb E\big[g'(h_i)\,g'(h_j)\big]\Big|_{W^*}=u_iu_j\,J_2(ij)\big|_{W^*},
\end{aligned}
\end{equation}

This new Lyapunov equation can now be solved numerically for $\tilde C\in\mathbb R^{K\times K}$. This allows for a very efficient way to compute the theoretical generalization error for a given choice of $\boldsymbol v$, $\boldsymbol u$, and hyperparameters:

\begin{equation}
    \epsilon_i=\operatorname{tr}(P_i\tilde C).
\end{equation}

As seen from Figure \ref{fig:2}, the theory (dotted line) matches the simulations and ODEs very well. In Appendix \ref{app:high-order}, we discuss the regimes of $\mu$ for which this approximation holds. Let us now return to our empirical observations.

\paragraph{Explaining Observation 1.} When $\alpha=\beta$,  a diagonal $\tilde C$ solve our Lyapunov equation exactly. It immediately follows that

\begin{equation}\ \alpha=\beta\ \Longrightarrow\ \epsilon_{\rm main}=\epsilon_{\rm aux}\end{equation}

for any activation and any correlation $\rho$. Substituting this result into Observation 2, we get

\begin{equation}\epsilon_{\rm main}|_{\alpha=\beta}=\alpha\epsilon_0.\end{equation}

\paragraph{Explaining Observation 2.}

Taking the trace of Equation \ref{eq:lyap_P}, we get

\begin{equation}\alpha\epsilon_{\text{main}}+\beta\epsilon_{\text{aux}}=\frac{\mu\sigma^2}{2}\,(G_\text{main}\alpha^2+G_\text{aux}\beta^2)\end{equation}

where $G_\text{main} = \operatorname{tr}P_\text{main}=\sum_i v_i^2\,J_2(i,i)|_{W^*}$, and similarly for $G_\text{aux}$.

At the fixed point $W^*$, $Q_{ii}\to T_{ii}=1$ (the student aligns with the teacher), so one can compute directly $J_2(i, i)|_{W^*}=\mathbb E[g'(h)^2]=J_2^*$. Then it immediately follows, since the readouts have the same norm, that $G_\text{main}=G_\text{aux}=G=J_2^*|v|^2$, and

\begin{equation}\alpha\epsilon_{\text{main}}+\beta\epsilon_{\text{aux}}=\frac{\mu\sigma^2 G}{2}\,(\alpha^2+\beta^2)=\epsilon_{0}\,(\alpha^2+\beta^2),\end{equation}

with the auxiliary-free floor $\epsilon_{0}$ obtained by setting $\alpha=1,\beta=0$. This result holds to leading order in $\mu$. In Appendix \ref{app:high-order}, we characterize the impact of higher-order corrections, showing agreement for $\mu \leq 10^{-2}$ and increasing deviations for $\mu\to1$.

We can compute $G$ analytically from $J_2$, and we get 

\begin{equation}
\begin{aligned}
G^\text{linear}=|\boldsymbol v|^2,\qquad
G^{\text{erf}}=\dfrac{2}{\pi\sqrt3}|\boldsymbol v|^2,\qquad G^{\text{ReLU}}=\dfrac12|\boldsymbol v|^2.
\end{aligned}
\end{equation}

From this, we can recover the generalization error of the online single-output teacher-student for any activation function, namely, from

\begin{equation}
\epsilon_0=\frac{\mu\sigma^2 G}{2},
\end{equation}

which aligns with our previous results for linear models, and generalizes previously known results from \citet{goldt_2020}, which proposed an approach that requires the computation of all integrals in Equation \ref{eq:integrals} of Appendix \ref{app:diff_eq} to obtain the leading term in $\sigma^2$ of the generalization error. Our approach requires only $J_2$, which is much easier to compute, but is limited to the leading term in $\mu$.

\begin{figure*}[!ht]
    \centering
    \includegraphics[width=\linewidth]{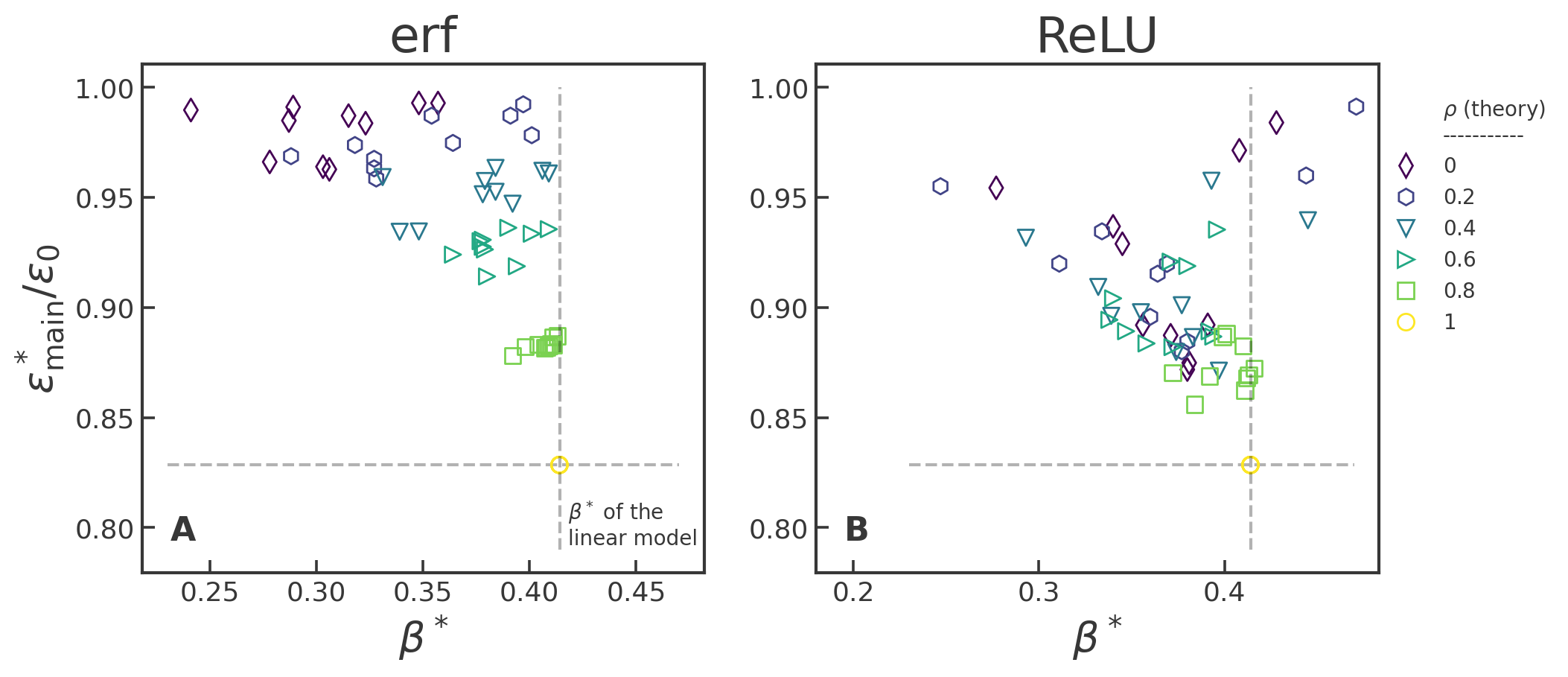}
    \caption{\textbf{Best improvement (theory).} Best relative loss against optimal $\beta^*$ computed for the model with A) erf and B) ReLU activation, $\alpha=1$. We uniformly sample ten $\boldsymbol v$, $\boldsymbol u$ with norm $|\boldsymbol v|=|\boldsymbol u|=\sqrt K$ and given correlation $\rho$. Results obtained by numerically solving the corresponding Lyapunov equation for $\mu=0.01$, $\sigma=0.1$, $K=4$.}
    \label{fig:improv_theo}
\end{figure*}

\paragraph{Explaining Observation 3.} 

In Figure \ref{fig:improv_theo}, we randomly sample $\boldsymbol v$ and $\boldsymbol u$ with the same norm and a given correlation $\rho$, and we numerically solve the corresponding Lyapunov equation for different $\beta$s. From this, we compute the optimal $\beta^*$ and the relative loss $\epsilon_\text{main}^*/\epsilon_0\equiv \epsilon_\text{main}(\beta^*)/\epsilon_0$, and we observe that all of the $\beta^*$s lie below $\sqrt 2 - 1$ for models with erf activation, but not for ReLU.

In Appendix \ref{app:beta_star}, we discuss the case where $P_\text{main}$ and $P_\text{aux}$ commute, and we show that, under this strong assumption, which does not hold generally, $\beta^*<1/2$ when $\alpha=1$.

\section{Discussion and Future Work}

We showed that auxiliary learning improves performance for the teacher-student setup in the large-input and small-learning-rate limit. We can write the linear model's generalization error analytically, and it exhibits an interesting phenomenology. Namely, the optimal $\beta^*$ weighting the auxiliary task is always $\sqrt2-1$ regardless of the correlation between tasks, and the generalization errors of the main and auxiliary tasks are tied to the generalization of the single-output model by an elegant relationship (Equation \ref{eq:loss_relationship}).

We then made similar empirical observations on models with nonlinear activations. The optimal $\beta^*$ is not always the same as the linear model; the advantage curves remain similar, and the loss relationship stays empirically valid. This led us to develop an analytic theory by expanding online SGD around the teacher optimum $W^*$. We obtain a Lyapunov equation that we can solve numerically to predict the generalization error for any choice of second-layer weights, activation function, and hyperparameters. This theory also explains our empirical observations and could be trivially expanded to more complicated setups, such as multiple auxiliary outputs or tasks with different noise scales, even though we explored these setups only for the linear model. One of the most relevant consequences is that we can now predict the first-order term in $\mu$ of the generalization error for a one-output teacher-student with a fixed second layer and any nonlinear activation, a novel result in the teacher-student literature.

A future direction left unexplored is student models with a different number of neurons from the teacher. \citet{goldt_2020} highlighted an empirical difference in behavior between sigmoids and ReLUs, which our analytical theory could explain. Another interesting direction is to extend our results to networks with learnable second-layer weights. Once again, the model convergence can be studied using our fluctuation-dissipation argument, although the application would not be trivial since the learning rate effectively varies between student weights. Lastly, it should be possible to generalize our results to non-Gaussian inputs using Gaussian equivalence theory \citep{goldt_2022}.

\subsubsection*{Acknowledgements}

\textit{Funding:} The Ph.D. fellowship of F. Milanesio is cofinanced by the company \textit{Additati\&Partners Consulting s.r.l.} at UniTO.

\subsubsection*{Contributions}

MO conceived the original idea. FM developed the theory with assistance from AI (the researcher, not a language model) and wrote the code for the numerical simulations. All authors revised the final manuscript.

\subsubsection*{AI use statement}

In this work, LLMs provided an important link to existing literature, namely with dissipation-fluctuations theory and adaptive filters, which were critical ingredients in developing our mathematical framework. After an initial formulation of generic mathematical claims, their application in the context of our work was human-made. We did not use generative AI tools to propose or refine hypotheses, design or provide feedback on research methodology or experiments, implement methods, assist with translation, support qualitative and thematic data analysis, or interpret results. Dataset-related uses do not apply to this work. Additionally, we used generative AI tools as an aid in writing software code, while most of our code is human-written, and to review the manuscript. We have reviewed all AI-assisted work. We take responsibility for the final content of this work, including text, claims, or artifacts produced with the aid of LLMs.

\subsubsection*{Reproducibility statement}

Code available to reproduce the numerical simulations at:

\href{https://github.com/FedericoMilanesio/teacher-student}{https://github.com/FedericoMilanesio/teacher-student}.

\bibliographystyle{iclr2026_conference}
\bibliography{references} 

\newpage

\renewcommand{\thefigure}{S\arabic{figure}}
\setcounter{figure}{0}
\renewcommand{\thetable}{S\arabic{table}}
\setcounter{table}{0}

\appendix

\section*{Appendix} 

\section{Derivation of the auxiliary dynamics differential equations}\label{app:diff_eq}
The derivation of the set of differential equations for the order parameters closely follows the derivation for the one-output case by \citet{saad_1995a, saad_1995b, biehl_1995, vicente_1997, saad_1999, inoue_2003, goldt_2020}. Let us consider the generalization error

\begin{equation}
    \epsilon_{\text{main}} (\theta, \theta^*)= \underset{\boldsymbol x\in \mathcal{N}}{\mathbb{E}}\bigg[\bigg( y^*_{\text{main}} (\boldsymbol{x})-y_{\text{main}} (\boldsymbol{x})\bigg)^2\bigg].
\end{equation}

Expanding the square, we immediately get

\begin{equation}
\begin{aligned}
    \epsilon_{\text{main}} &= \sum_{i,j}v_iv_j\mathbb{E}\left[g(h_i)g(h_j)\right]+\sum_{n,m}v_n^*v_m^*\mathbb{E}\left[g(h_n^*)g(h_m^*)\right]-2\sum_{i,n}v_iv_n^*\mathbb{E}\left[g(h_i)g(h_n^*)\right]\\&= \sum_{i,j}v_iv_jI_2(i,j)+\sum_{n,m}v_n^*v_m^*I_2(n,m)-2\sum_{i,n}v_iv_n^*I_2(i,n),
\end{aligned}
\end{equation}

with the notation $I_2(i,j)=\mathbb{E}[g(h_i)\,g(h_j)]$, $i$, $j$ being indices of units in the student network and $n$, $m$ teacher ones. Since the inputs are taken to be $i.i.d.$ Gaussian, the pre-activations 

\begin{equation}
\label{pre-act}
\begin{aligned}
    h_i =\dfrac{\boldsymbol w_i\cdot\boldsymbol x}{\sqrt N}
\end{aligned}
\end{equation}

are jointly Gaussian together with the teacher pre-activations $\boldsymbol h^*$, having zero-mean and a $2K \times 2K$ covariance matrix with block structure

\begin{equation}
\begin{aligned}
    \text{cov}=\begin{pmatrix}
        Q & R\\
        R^T & T
    \end{pmatrix},
\end{aligned}
\end{equation}

with 

\begin{equation}
\begin{aligned}
Q_{ik} &= \mathbb{E}\left[h_ih_k\right]=\dfrac{\boldsymbol w_i^T}{\sqrt N}\mathbb{E}\left[\boldsymbol x \boldsymbol x^T\right]\dfrac{\boldsymbol w_k }{\sqrt N}=\dfrac{\boldsymbol w_i^T \boldsymbol w_k}{N},\\R_{in} &=\mathbb{E}\left[h_ih_n^*\right]= \dfrac{\boldsymbol w_i^T \boldsymbol w_n^*}{N},\\\,T_{mn} &=\mathbb{E}\left[h^*_mh^*_n\right]= \dfrac{\boldsymbol w_m^{*T} \boldsymbol w_n^*}{N}\xrightarrow[N \to \infty]{}\delta_{mn}.
\end{aligned}
\end{equation}

Notice that $Q$ and $R$ evolve during training, while $T$ does not.

To compute the generalization error, we only need to compute either analytically or numerically the integral $I_2$ (namely, a two-dimensional Gaussian
average of a nonlinear function). Notice the convention, valid from now on, that $i$, $k$ (and $j$, $l$) denote student indices, and $n$, $m$ teacher indices.

Let us now consider the dynamics. Starting from Equation \ref{eq:sgd}, for the two-output case, they are described by the set of equations

\begin{equation}
\label{eq:sgd2}
\begin{aligned}
    W^{t+1}_{ik}&= W^t_{ik}-\mu\Bigg[\alpha\,v_i\Bigg(\sum_m v_m^*\,g\left(\dfrac{\boldsymbol w_m^*\cdot\boldsymbol x}{\sqrt N}\right)-\sum_j v_j\,g\left(\dfrac{\boldsymbol w_j\cdot\boldsymbol x}{\sqrt N}\right)+\sigma z_{\text{main},t}\Bigg)\\ &+\beta\, u_i\Bigg(\sum_m u_m^*\,g\left(\dfrac{\boldsymbol w_m^*\cdot\boldsymbol x}{\sqrt N}\right)-\sum_j u_j\,g\left(\dfrac{\boldsymbol w_j\cdot\boldsymbol x}{\sqrt N}\right)+\sigma z_{\text{aux},t}\Bigg)\Bigg]g'\left(\dfrac{\boldsymbol w_i\cdot\boldsymbol x}{\sqrt N}\right)\dfrac{x_k}{\sqrt N},\\
    v^{t+1}_i&= v^t_i-\dfrac{\mu}{N}\alpha \Bigg(\sum_m v_m^*\,g\left(\dfrac{\boldsymbol w_m^*\cdot\boldsymbol x}{\sqrt N}\right)-\sum_j v_j\,g\left(\dfrac{\boldsymbol w_j\cdot\boldsymbol x}{\sqrt N}\right)+\sigma z_{\text{main},t}\Bigg)\,g\left(\dfrac{\boldsymbol w_i\cdot\boldsymbol x}{\sqrt N}\right),\\
    u^{t+1}_i&= u^t_i-\dfrac{\mu}{N}\beta \Bigg(\sum_m u_m^*\,g\left(\dfrac{\boldsymbol w_m^*\cdot\boldsymbol x}{\sqrt N}\right)-\sum_j u_j\,g\left(\dfrac{\boldsymbol w_j\cdot\boldsymbol x}{\sqrt N}\right)+\sigma z_{\text{aux},t}\Bigg)\,g\left(\dfrac{\boldsymbol w_i\cdot\boldsymbol x}{\sqrt N}\right),
\end{aligned}
\end{equation}

where the second-layer weights have a rescaled learning rate $\mu/N$. Taking the appropriate expectation of these equations, and rescaling the time variable $\tau = t/N$,  we obtain

\begin{subequations}
\begin{align}
\odv{v_k}{\tau} &= \alpha\mu\bigg(\sum_m v_m^*I_2(k,m)-\sum_j v_jI_2(k,j) \bigg),\\
\odv{u_k}{\tau} &= \beta\mu\bigg(\sum_m u_m^*I_2(k,m)-\sum_j u_jI_2(k,j)\bigg),\\
\odv{Q^{ik}}{\tau}&=\mu\sum_m \Big[\left(\alpha v_m^*v_k+\beta u_m^*u_k\right)I_3(k,i,m)+\left(\alpha v_m^*v_i+\beta u_m^*u_i\right)I_3(i,k,m)\Big] \\&-\mu\sum_j \Big[\left(\alpha v_jv_k+\beta u_ju_k\right)I_3(k,i,j)\nonumber+\left(\alpha v_jv_i+\beta u_ju_i\right)I_3(i,k,j)\Big]\nonumber\\
&+\mu^2\sum_{n,m}\Big[\left(\alpha^2v_iv_kv_m^*v_n^*+\alpha\beta(v_ku_i+v_iu_k)u_n^*v_m^*+\beta^2u_iu_ku_n^*u_m^*\right)I_4(i,k,n,m)\Big]\nonumber\\
&+\mu^2\sum_{j,l}\Big[\left(\alpha^2 v_iv_kv_jv_l+\alpha\beta(v_ku_i+v_iu_k)u_jv_l+\beta^2u_iu_ku_ju_l\right)I_4(i,k,j,l)\Big]\nonumber\\
&-\mu^2\sum_{j,n}\Big[\left(2\alpha^2 v_iv_kv_jv_n^*+\alpha\beta(v_ku_i+v_iu_k)(u_jv_n^*+v_ju_n^*)+2\beta^2u_iu_ku_ju_n^*\right)I_4(i,k,j,n)\Big]\nonumber\\
&+\mu^2\sigma^2(\alpha^2v_kv_i+\beta^2u_ku_i)J_2(i,k)\nonumber,\\
\odv{R^{in}}{\tau}&=\mu\sum_m \left(\alpha v_m^*v_i+\beta u_m^*u_i\right)I_3(i,n,m)-\mu\sum_j\left(\alpha v_iv_j+\beta u_iu_j\right)I_3(i,n,j),
\end{align}
\end{subequations}

depending on the integrals

\begin{equation}
\label{eq:integrals}
\begin{aligned}
I_2 &= \mathbb{E}[g(x)\,g(y)], \\
I_3 &= \mathbb{E}[g'(x)\,y\,g(z)], \\
I_4 &= \mathbb{E}[g'(x)\,g'(y)\,g(z)\,g(h)], \\
J_2 &= \mathbb{E}[g'(x)\,g'(y)].
\end{aligned}
\end{equation}

Thus, the system dynamics are described by a non-extensive set of mean-field ODEs, which can be numerically integrated as long as we can compute these integrals. We will now discuss two common choices of activation functions.

\subsection{Error function}

If $g(x)= \text{erf}(x/\sqrt2)$, then all integrals can be computed analytically \citep{saad_1995a, saad_1995b, biehl_1995, goldt_2020} and we get

\begin{equation}
\begin{aligned}
I_2(x,y)&=\mathbb{E}\left[ g(x)g(y)\right]=\frac{2}{\pi}\arcsin\left(\frac{c_{xy}}{\sqrt{(1+c_{xx})(1+c_{yy})}}\right),\\
J_2(x,y)&=\mathbb{E}\left[ g'(x)g'(y)\right]=\frac{2}{\pi \sqrt{\Lambda_4}},\\
I_3(x,y,z)&=\mathbb{E}\left[ g'(x)\,y\,g(z)\right]=\frac{2}{\pi}\frac{(1+c_{xx})c_{yz}-c_{yx}c_{xz}}
{(1+c_{xx})\sqrt{\Lambda_4}},\\
I_4(x,y,z,h)&=\mathbb{E}\left[ g'(x)g'(y)g(z)g(h)\right]=\frac{4}{\pi^2}
\frac{ 1 }
{\sqrt{
\Lambda_4
} }\arcsin\Big( \dfrac{\Lambda_0}{\sqrt{\Lambda_1\Lambda_2}} \Big),
\end{aligned}
\end{equation}

with 

\begin{equation}
\begin{aligned}
\Lambda_4 &= (1+c_{xx})(1+c_{yy}) - c_{xy}^2,\\
\Lambda_0 & =\Lambda_4c_{zh} - c_{yz}c_{yh}(1+c_{xx}) - c_{xz}c_{xh}(1+c_{yy}) + c_{xy}(c_{xz}c_{yh}+c_{yz}c_{xh}),\\
\Lambda_1&= \Lambda_4 (1 + c_{zz}) - c_{yz}^2(1+c_{xx}) - c_{xz}^2(1+c_{yy}) + 2c_{xy}c_{xz}c_{yz},\\
\Lambda_2 &= \Lambda_4 (1 + c_{hh}) - c_{yh}^2(1+c_{xx}) - c_{xh}^2(1+c_{yy}) + 2c_{xy}c_{xh}c_{yh}.
\end{aligned}
\end{equation}

\subsection{Rectified linear unit} \label{app:relu}

If $g(x)= \text{ReLU}(x)=\max(x,0)$, then some of the integrals can be computed analytically \citep{straat_2019}, namely 

\begin{equation}
\begin{aligned}
I_2(x,y)&=\mathbb{E}\left[ g(x)g(y)\right]=\dfrac{c_{xy}}{4}+\dfrac{1}{2\pi}\left(c_{xy}\,\text{arcsin}\left(\dfrac{c_{xy}}{\sqrt{c_{xx}c_{yy}}}\right)+\sqrt{c_{xx}c_{yy}-c_{xy}^2}\right),\\
J_2(x,y)&=\mathbb{E}\left[ g'(x)g'(y)\right]=\dfrac{1}{4}+\dfrac{1}{2\pi}\text{arcsin}\left(\dfrac{c_{xy}}{\sqrt{c_{xx}c_{yy}}}\right),\\
I_3(x,y,z)&=\mathbb{E}\left[ g'(x)\,y\,g(z)\right]=\dfrac{c_{xy}}{2\pi c_{xx}}\sqrt{c_{xx}c_{zz}-c_{xz}^2} + \dfrac{1}{2\pi}c_{yz}\text{arcsin}\left(\dfrac{c_{xz}}{\sqrt{c_{xx}c_{zz}}}\right) + \dfrac{c_{yz}}{4}.
\end{aligned}
\end{equation}

The integral $I_4=\mathbb{E}\left[ g'(x)g'(y)g(z)g(h)\right]=\mathbb{E}\left[ \mathbf{1}_{x>0}\,\mathbf{1}_{y>0}\,\text{max}(z,0)\,\text{max}(h,0)\right]$ cannot be computed analytically. \citet{citton_2024, citton_2025, citton_2025b} proposed a Hermite expansion to approximate it numerically, but we found that it suffers from convergence issues for ill-conditioned matrices. We instead used a conditional Gaussian decomposition with Gauss–Hermite quadrature to approximate it.

Partition $C$ as

\begin{equation}
C = \begin{pmatrix} A & B^T \\ B & D \end{pmatrix}, \quad A = \operatorname{Cov}(x,y),\quad D = \operatorname{Cov}(z,h),\quad B = \operatorname{Cov}\!\bigl((z,h),(x,y)\bigr).
\end{equation}

By the Gaussian conditioning formula, $(z,h)$ conditioned on $(x,y)=\mathbf{u}$ is Gaussian:

\begin{equation}
(z,h)\mid(x,y)=\mathbf{u} \;\sim\; \mathcal{N}\!\left(\underbrace{BA^{-1}\mathbf{u}}_{=\,\mathbf{m}(\mathbf{u})},\; \underbrace{D - BA^{-1}B^T}_{=\,S}\right).
\end{equation}

$S$ is the Schur complement of $A$ in $C$, and is positive semi-definite. Then our integral becomes

\begin{equation}
I_4 = \int_{\mathbb R_+^2}\underbrace{\mathbb{E}\!\left[\max(z,0)\,\max(h,0)\;\Big|\;(x,y)=\mathbf{u}\right]}_{=\;K(\mathbf{m}(\mathbf{u}),\, S)}\;p(\mathbf{u})\,d\mathbf{u},
\end{equation}

Writing $s_1=\sqrt{S_{11}}$, $s_2=\sqrt{S_{22}}$, $\rho = S_{12}/(s_1 s_2)$, $a=m_{1}/s_1$, $b=m_{2}/s_2$, $k=\sqrt{1-\rho^2}$, the closed form for $K$ is given by \citet{muthen_1990, kan_2017}.

\begin{equation}
K(\mathbf m, S) = s_1 s_2\Big[(ab+\rho)\,\Phi_2(a,b;\rho) + a\,\varphi(b)\,\Phi\!\Big(\tfrac{a-\rho b}{k}\Big) + b\,\varphi(a)\,\Phi\!\Big(\tfrac{b-\rho a}{k}\Big) + k^2\,\varphi_2(a,b;\rho)\Big],
\end{equation}

where $\varphi,\Phi$ are the univariate standard normal density and CDF, and $\varphi_2,\Phi_2$ their bivariate analogues at correlation $\rho$. The bottleneck in this expression is $\Phi_2$, which has no elementary closed form. We evaluate it via Owen's identity \citep{owen_1956},

\begin{equation}
\frac{\partial \Phi_2}{\partial \rho}(a,b;\rho) = \varphi_2(a,b;\rho) \quad\Longrightarrow\quad \Phi_2(a,b;\rho) = \Phi(a)\Phi(b) + \int_0^{\rho}\varphi_2(a,b;t)\,dt,
\end{equation}

which is a smooth, one-dimensional integral over a compact interval $[0,\rho]\subset[-1,1]$ and converges rapidly under low-order Gauss–Legendre quadrature.
Then $I_4$ can be approximated by Gauss–Hermite quadrature \citep{abramowitz_1964}. When $C$ is singular, then $I_4$ always reduces to one of two integrals that are known analytically. \citet{citton_2025b} described them in their additional material.

\section{Generalization error of the two-output linear model}\label{app:linear}

Given matrix $W^*\in\mathbb{R}^{K\times N}$ and two vectors $\boldsymbol v,\,\boldsymbol u\in\mathbb{R}^{K}$, we want to compute

\begin{equation}
    \epsilon_{\text{main}} = \lim_{t\to\infty}\epsilon_{\text{main}}(t) = \lim_{t\to\infty}\underset{x\sim\mathcal N}{\mathbb{E}}\left[\left(\dfrac{\boldsymbol{v}^TW^*\boldsymbol{x}}{\sqrt N}-\dfrac{\boldsymbol{v}^TW_t\boldsymbol{x}}{\sqrt N}\right)^2\right],
\end{equation}

where

\begin{equation}
    W_{t+1} = W_{t} - \mu \nabla_W \mathcal{L}(t),
\end{equation}

and

\begin{equation}
\begin{aligned}
     \mathcal{L}(t) =& \dfrac{1}{2}\Bigg[\alpha\left(\dfrac{\boldsymbol{v}^TW^*\boldsymbol{x}_t}{\sqrt N}-\dfrac{\boldsymbol{v}^TW_t\boldsymbol{x}_t}{\sqrt N}+\sigma z_{\text{main},t}\right)^2\\&+\beta\left(\dfrac{\boldsymbol{u}^TW^*\boldsymbol{x}_t}{\sqrt N}-\dfrac{\boldsymbol{u}^TW_t\boldsymbol{x}_t}{\sqrt N}+\sigma z_{\text{aux},t}\right)^2\Bigg],
    \end{aligned}
\end{equation}

with $\alpha, \beta \in \mathbb{R}^+,\,\alpha\neq 0$, and $\boldsymbol{x}_t\sim\mathcal{N}(0,I_N)$ and $z_{\text{main}},z_{\text{aux}}\in\mathcal{N}(0,1)$.

Then,

\begin{equation}
\begin{aligned}
    W_{t+1} = W_{t} + \mu\Bigg[&\alpha\left(\dfrac{\boldsymbol{v}^TW^*\boldsymbol{x}_t}{\sqrt N}-\dfrac{\boldsymbol{v}^TW_t\boldsymbol{x}_t}{\sqrt N}+\sigma z_{\text{main},t}\right)\boldsymbol{v}\boldsymbol{x}_t^T\\+&\beta\left(\dfrac{\boldsymbol{u}^TW^*\boldsymbol{x}_t}{\sqrt N}-\dfrac{\boldsymbol{u}^TW_t\boldsymbol{x}_t}{\sqrt N}+\sigma z_{\text{aux},t}\right)\boldsymbol{u}\boldsymbol{x}_t^T\Bigg].
    \end{aligned}
\end{equation}

We can now project the weight matrix onto the task-relevant directions, calling $\boldsymbol p_t:={\boldsymbol v_t^TW}/{\sqrt N}$ and $\boldsymbol q_t:={\boldsymbol u_t^TW}/{\sqrt N}$, then

\begin{equation}
\begin{aligned}
    \boldsymbol p_{t+1} = \boldsymbol p_{t} + \dfrac{\mu}{N}\bigg[&\alpha |\boldsymbol v|^2(\boldsymbol p^{*T}\boldsymbol{x}_t-\boldsymbol p_t^T\boldsymbol{x}_t+\sigma z_{\text{main},t})\boldsymbol{x}_t\\+&\beta \rho  |\boldsymbol v||\boldsymbol u| (\boldsymbol q^{*T}\boldsymbol{x}_t-\boldsymbol q_t^T\boldsymbol{x}_t+\sigma z_{\text{aux},t})\boldsymbol{x}_t\bigg],\\  \boldsymbol  q_{t+1} = \boldsymbol q_{t} + \dfrac{\mu}{N}\bigg[&\alpha \rho  |\boldsymbol v||\boldsymbol u|(\boldsymbol p^{*T}\boldsymbol{x}_t-\boldsymbol p_t^T\boldsymbol{x}_t+\sigma z_{\text{main},t})\boldsymbol{x}_t\\+&\beta |\boldsymbol u|^2(\boldsymbol q^{*T}\boldsymbol{x}_t-\boldsymbol q_t^T\boldsymbol{x}_t+\sigma z_{\text{aux},t})\boldsymbol{x}_t\bigg],
\end{aligned}
\end{equation}

where $\rho=\boldsymbol u^T \boldsymbol v / |\boldsymbol v||\boldsymbol u|$.

Introducing the deviations from the teacher $\delta{\boldsymbol p}_t = {\boldsymbol p}^*-{\boldsymbol p}_t$ and $\delta{\boldsymbol q}_t = {\boldsymbol q}^*-{\boldsymbol q}_t$, we get

\begin{equation}
\begin{aligned}
\label{eq:pq_dynamics}
    \delta{\boldsymbol p}_{t+1} = \delta{\boldsymbol p}_{t}  -\dfrac{\mu}{N}\left[\alpha |\boldsymbol v|^2(\delta{\boldsymbol p}^{T}_t\boldsymbol{x}_t+\sigma z_{\text{main},t})\boldsymbol{x}_t+\beta \rho  |\boldsymbol v||\boldsymbol u| (\delta{\boldsymbol q}^{T}_t\boldsymbol{x}_t+\sigma z_{\text{aux},t})\boldsymbol{x}_t\right],\\  \delta{\boldsymbol q}_{t+1} = \delta{\boldsymbol q}_{t} - \dfrac{\mu}{N}\left[\alpha \rho  |\boldsymbol v||\boldsymbol u|(\delta{\boldsymbol p}^{T}_t\boldsymbol{x}_t+\sigma z_{\text{main},t})\boldsymbol{x}_t+\beta |\boldsymbol u|^2(\delta{\boldsymbol q}^{T}_t\boldsymbol{x}_t+\sigma z_{\text{aux},t})\boldsymbol{x}_t\right].
\end{aligned}
\end{equation}

Then, the generalization error at time $t$ can be computed as

\begin{equation}
    \epsilon_{\text{main}}(t) =\mathbb{E}_x\left[(\delta{\boldsymbol{p}}^T_t\boldsymbol{x})^2\right]=\delta{\boldsymbol{p}}^T_t\delta{\boldsymbol{p}}_t.
\end{equation}

We introduce the variable $\boldsymbol\phi_t^T = (\delta{\boldsymbol{p}}^T_t,\delta{\boldsymbol{q}}^T_t)$, and we choose, for simplicity, our second layer weights to have the same norm, namely $|\boldsymbol v|=|\boldsymbol u|$. Then,

\begin{equation}
    \boldsymbol\phi_{t+1} = \left(I-\eta A\otimes(\boldsymbol x_t\boldsymbol x^T_t)\right) \boldsymbol\phi_{t}-\eta\sigma\,\boldsymbol b_t \otimes  \boldsymbol x_t,
\end{equation}

where we defined a new learning rate $\eta := \dfrac{\mu |\boldsymbol v|^2}{N}$, as well as the quantities

\begin{equation}
   A = \begin{pmatrix}\alpha & \beta \rho\\ \alpha \rho & \beta\end{pmatrix},
\end{equation}

and

\begin{equation}
   \boldsymbol b_t = \begin{pmatrix}\alpha \,z_{\text{main},t} + \beta \rho\,z_{\text{aux},t}\\ \alpha \rho \,z_{\text{main},t}+ \beta\,z_{\text{aux},t}\end{pmatrix}.
\end{equation}

Given that the weights at time $t$ are independent of the current input $\boldsymbol x_t$, namely $\boldsymbol x_t \perp \boldsymbol\phi_t$, then it follows, taking the expectation over the whole dynamics

\begin{equation}
   \mathbb{E}[ \boldsymbol\phi_{t+1}] = \left(I-\eta A\otimes I\right) \mathbb{E}[ \boldsymbol\phi_{t}],
\end{equation}

which converges if 

\begin{equation}
    0<\mu<\dfrac{2N}{|\boldsymbol v|^2\,\lambda_{max}(A)}= \dfrac{4N}{|\boldsymbol v|^2\left(\alpha+\beta + \sqrt{(\alpha-\beta)^2 + 4\rho^2\alpha\beta}\right)}.
\end{equation}

We are then interested in the quantity $C_t= \boldsymbol\phi_{t}\boldsymbol\phi_{t}^T$, from which we can derive the generalization error. 

Applying the Gaussian moment theorem results in the identity 

\begin{equation}\mathbb E[\boldsymbol x \boldsymbol x^T M \boldsymbol x \boldsymbol x^T]
=\operatorname{tr}(M) I + M + M^T,\end{equation}

we get

\begin{equation}
\begin{aligned}
C_{t+1}&=  C_{t}-\eta (A\otimes I) C_{t} -\eta C_{t}(A^T\otimes I) \\&+\eta^2\sigma^2\Sigma\otimes I+\mathcal O (\eta^3),
\end{aligned}
\end{equation}

where

\begin{equation}
   \Sigma =\mathbb{E}_t\bigg[ \boldsymbol b_t\boldsymbol b_t^T\bigg]= \begin{pmatrix}\alpha^2+\beta^2\rho^2 & \rho(\alpha^2+\beta^2)\\ \rho(\alpha^2+\beta^2) & \alpha^2\rho^2+\beta^2\end{pmatrix}=AA^T.
\end{equation}

Our matrix $C$ inherits the Kronecker structure from the other terms, namely $C = \tilde C\otimes I$. This reduces the equation to

\begin{equation}
   \tilde C_{t+1}=  \tilde C_{t}-\eta A\tilde C_{t} -\eta\, \tilde C_{t}A^T+\eta^2 \sigma^2\Sigma+\mathcal O (\eta^3),
\end{equation}

Now, we impose stationary, setting $\tilde C_{t+1}=\tilde C_t=\tilde C_\infty$, and we can solve for $\tilde C_\infty$, obtaining

\begin{equation}
\begin{aligned}
    \epsilon_{\text{main}} &= \lim_{t\to\infty}\big[\epsilon_{\text{main}}(t)\big] = \delta{\boldsymbol{p}}^T_\infty\delta{\boldsymbol{p}}_\infty= \sum_{i=1}^N(C_\infty)_{ii}=N (\tilde C_\infty)_{11}=\\&=\dfrac{\mu \sigma^2|\boldsymbol v|^2}{2}\dfrac{\alpha^2+\alpha\beta(1-\rho^2)+\beta^2\rho^2}{\alpha+\beta}+\mathcal{O}(\mu^2).
    \end{aligned}
\end{equation}

\paragraph{Different noise variance in auxiliary labels.} Let us now consider the same linear model as before, with the addition that the auxiliary labels $y_{\text{aux}}$ may have noise with a different variance $\sigma_{\text{aux}}^2$, while the main task noise has variance $\sigma^2$. With an analogous calculation, the leading order in $\mu$ of the generalization error is

\begin{equation}
\begin{aligned}
\label{err_linear_noisy}
    \epsilon_{\text{main}} &= \dfrac{\mu|\boldsymbol v|^2\sigma^2}{2}\dfrac{\left(\alpha^2+\alpha\beta(1-\rho^2)\right)+\dfrac{\sigma_{\text{aux}}^2}{\sigma^2}\rho^2\beta^2}{\alpha+\beta}=\\
    &=\epsilon_{0} \dfrac{\alpha^2+\alpha\beta(1-\rho^2)+{\delta^2}\rho^2\beta^2}{\alpha+\beta},
\end{aligned}
\end{equation}

where we have defined the noise scale $\delta={\sigma_{\text{aux}}}/{\sigma}$.

\begin{figure*}[!ht]
    \centering
    \includegraphics[width=0.45\linewidth]{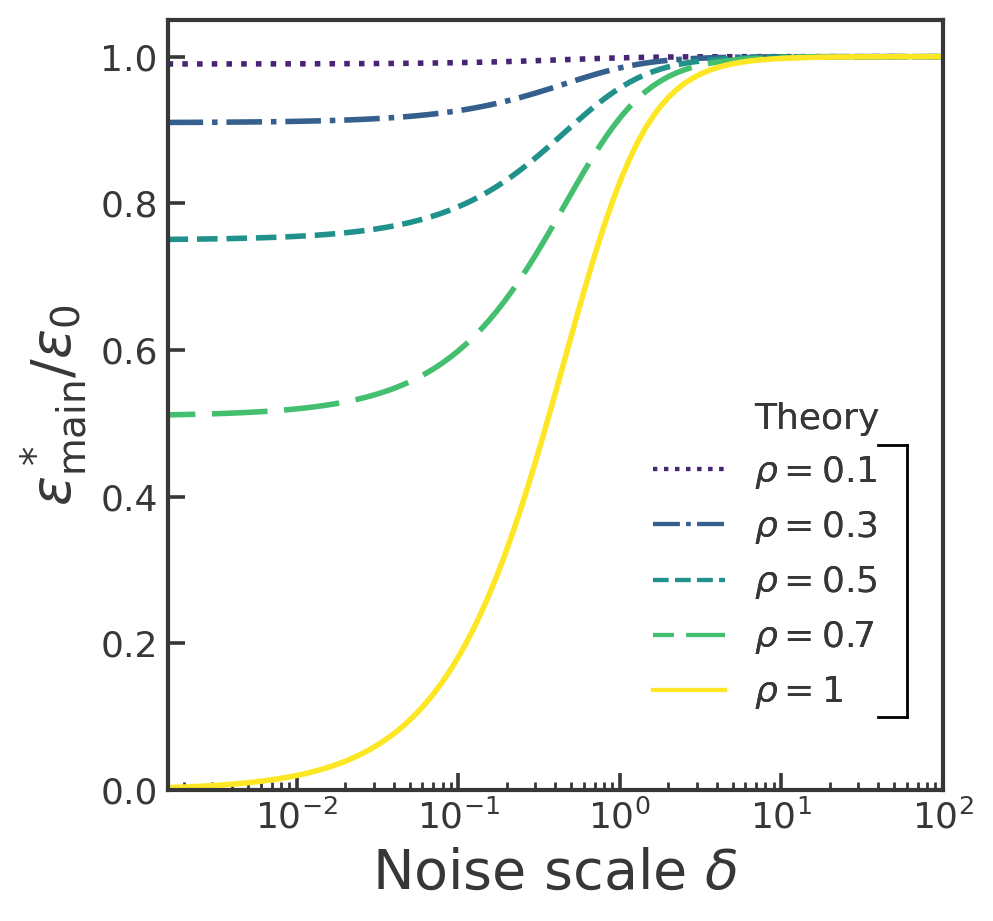}
    \caption{\textbf{Optimal generalization for different noise scales.} Leading term of the generalization error on the main task of the two-output linear model computed at the optimal $\beta^*$ as a function of noise scale $\delta=\sigma_\text{aux}/\sigma$ and different $\rho$, with $\alpha = 1$.}
    \label{fig:2_bis}
\end{figure*}

When $\alpha=1$

\begin{equation}
    \epsilon_{\text{main}}\Big|_{\alpha=1}=\epsilon_0\left(1-\rho^2\dfrac{\beta(1-\beta\delta^2)}{1+\beta}\right)
\end{equation}

minimized by

\begin{equation}
    \beta^*={\sqrt{1+\dfrac{1}{\delta^2}}}-1.
\end{equation}

Notice that $\beta^*$ can be larger than $1$ and does not depend on $\rho$. For every choice of $\sigma_{\text{aux}}^2$, including when $\sigma_{\text{aux}}^2>\sigma^2$, there is always a $\beta^*\neq 0$ that improves the corresponding generalization error $\epsilon_{\text{main}}|_{\beta^*}=\epsilon_{\text{main}}^*$ compared to $\epsilon_0$, as shown in Figure \ref{fig:2_bis}. Of course, when $\delta>1$, the optimal $\beta^*$ is small with only marginal improvements in generalization, and setting $\beta>\beta^*$ can degrade performance. 

\paragraph{Multiple auxiliary labels.} Consider a model with two auxiliary labels instead $y_{\text{aux},1}$ and $y_{\text{aux},2}$, each having the same label noise $\sigma$. Namely, the training loss will be

\begin{equation}
\begin{aligned}
    \mathcal{L}(\boldsymbol x)&=\alpha\mathcal{L}_{\text{main}}(\boldsymbol x)+\beta\mathcal{L}_{\text{aux},1}(\boldsymbol x)+\gamma\mathcal{L}_{\text{aux},2}(\boldsymbol x),\\\mathcal{L}_i(\boldsymbol x)&=\frac{1}{2}\bigg( y_i ^*(\boldsymbol{x})-y_i (\boldsymbol{x})+\sigma z_i\bigg)^2.
\end{aligned}
\end{equation}

From this, the generalization error on the main task is, at the first order in $\mu$,

\newcommand{\Sbg}{\beta(1-\rho_{1}^{2})+\gamma(1-\rho_{2}^{2})}

\begin{equation}
\epsilon_{\mathrm{main}}
  = \frac{\mu\,\sigma^{2}\,\lvert\boldsymbol{v}\rvert^{2}}{2}\,
    \frac{\mathcal{N}}{\mathcal{D}} ,
\end{equation}

with numerator

\begin{equation}
\begin{aligned}
\mathcal{N}
&= \alpha^{3}\biggl(\Sbg\biggr) \\
&\quad{}+ \alpha^{2}\Bigl(
      \bigl[\Sbg\bigr]^{2}
      + \beta\gamma\bigl[1-\rho_{12}^{2}-\Delta\bigr]
    \Bigr) \\
&\quad{}+ \alpha\Bigl(
      \beta^{3}\rho_{1}^{2}(1-\rho_{12}^{2})
      + \gamma^{3}\rho_{2}^{2}(1-\rho_{2}^{2})\\
&\qquad\quad{}+ \beta\gamma\bigl[
        (\beta+\gamma)\Delta
        + \beta\rho_{1}^{2}(1-\rho_{2}^{2})
        + \gamma\rho_{2}^{2}(1-\rho_{12}^{2})
      \bigr]
    \Bigr)\\
&\quad{}+ \beta\gamma\Bigl(
      (\beta^{2}\rho_{1}^{2}+\gamma^{2}\rho_{2}^{2})(1-\rho_{12}^{2})
      + \beta\gamma\bigl[
          (\rho_{1}-\rho_{2}\rho_{12})^{2}
          + (\rho_{2}-\rho_{1}\rho_{12})^{2}
        \bigr]
    \Bigr),
\end{aligned}
\end{equation}

and denominator

\begin{equation}
\mathcal{D}
  = (\alpha+\beta)(\alpha+\gamma)(\beta+\gamma)
    - (\alpha+\beta+\gamma)\bigl(
        \alpha\beta\rho_{1}^{2}+\alpha\gamma\rho_{2}^{2}+\beta\gamma\rho_{12}^{2}
      \bigr)
    + \alpha\beta\gamma(1-\Delta),
\end{equation}

where $\rho_{1}$ and $\rho_{2}$ are the correlations between the main task and the auxiliary tasks, $\rho_{12}$ is the correlation between the two auxiliary tasks, and $\Delta = 1 + 2\rho_{1}\rho_{2}\rho_{12} - \rho_{1}^2 - \rho_{2}^2 - \rho_{12}^2$ is the Gram determinant of the correlation matrix between second-layer weights

\begin{equation}
G = \begin{pmatrix}
1 & \rho_{1} & \rho_{2} \\
\rho_{1} & 1 & \rho_{12} \\
\rho_{2} & \rho_{12} & 1
\end{pmatrix}.
\end{equation}

One could focus on many different correlation patterns between tasks. For example, we choose two outputs uncorrelated with each other, each having the same correlation with the main task, namely $\rho_{12}=0$, $\rho_{1}=\rho_{2}=\rho$. We also choose $\gamma=\beta$, $\alpha=1$. Then

\begin{equation}
   \epsilon_{\text{main}}= \epsilon_{0}\left(1-2\rho^2\frac{ (1-\beta) \beta}{\beta +1}\right).
\end{equation}

It is straightforward to see that $\beta^*=\sqrt 2 -1$, analogous to the two-output setup, but corresponding to a better generalization error than the minimum of Equation \ref{eq:fixed_alpha_lin}.

Once again, regardless of the choices of correlation between outputs,

\begin{equation}
    \alpha\epsilon_{\text{main}}+\beta\epsilon_{\text{aux},1}+\gamma\epsilon_{\text{aux},2}= \epsilon_{0}\big(\alpha^2+\beta^2+\gamma^2\big).
\end{equation}

\section{Effective dynamics with perfectly correlated output}\label{app:comm}

Let us consider the case where the second layer weights are the same, $\boldsymbol v=\boldsymbol v^*=\boldsymbol u=\boldsymbol u^* $,

\begin{equation}
     \mathcal{L}(t) = \dfrac{1}{2}\bigg[\alpha(y^*_1-y_1+\sigma z_{\text{main},t})^2+\beta(y^*_1-y_1+\sigma z_{\text{aux},t})^2\bigg],
\end{equation}

and, regardless of the choice of activation $g$,

\begin{equation}
\begin{aligned}
     W^{t+1} &= W^t+\mu\bigg[\alpha(y^*_1-y_1+\sigma z_{\text{main},t})+\beta(y^*_1-y_1+\sigma z_{\text{aux},t})\bigg]\bigg(\boldsymbol v \odot\boldsymbol g'(h)\bigg)\dfrac{\boldsymbol x^T}{\sqrt N}=\\&=W^t+(\alpha+\beta)\mu\left[y^*_1-y_1+\sigma \dfrac{\alpha z_{\text{main},t}+\beta z_{\text{aux},t}}{\alpha+\beta}\right]\bigg (\boldsymbol v \odot\boldsymbol g'(h)\bigg)\dfrac{\boldsymbol x^T}{\sqrt N},
\end{aligned}
\end{equation}

where $\boldsymbol g'(h)_i\equiv g'(h_i)$ and $\odot$ is the Hadamard (element-wise) product. This is equivalent to training a teacher-student network with one output, with an effective learning rate and noise:

\begin{equation}
\begin{aligned}
     \mu_{\text{eff}}&=(\alpha+\beta)\mu,\\\sigma^2_{\text{eff}}&=\dfrac{\alpha^2+\beta^2}{(\alpha+\beta)^2}\sigma^2.
\end{aligned}
\label{eq:effective_lr}
\end{equation}

\section{Universality of loss}\label{app:universality}

\subsection{Derivation of the Lyapunov equation}\label{app:lyap}

Empirically, as discussed previously, for good initializations the dynamics converge near the realizable fixed point $W^*$ (where the student recovers the teacher weights). Let us assume the second-layer weights are fixed, and write the weight difference $\delta \boldsymbol W=\boldsymbol W-\boldsymbol W^*$, where $ \boldsymbol W=\text{vec}(W)$ for ease of notation. The online SGD update with a new sample is

\begin{equation}\boldsymbol W_{t+1}=\boldsymbol W_t-\mu\,\nabla_{\boldsymbol W}{\mathcal L}(W_t; \boldsymbol x_t, \boldsymbol z_t)\end{equation}

The sample gradient close to the optimum $\boldsymbol W^*$ can be linearized as

\begin{equation}
\begin{aligned}
    \nabla_{\boldsymbol W}{\mathcal L}(\boldsymbol W_t; \boldsymbol x_t, \boldsymbol z_t)&=\nabla_{\boldsymbol W}\mathcal L(\boldsymbol W^*; \boldsymbol x_t, \boldsymbol z_t)+H_t\,\delta \boldsymbol W_t+o(\|\delta \boldsymbol W\|),\\ \text{with}\; H_t:&=\nabla^2_{\boldsymbol W}\mathcal L(\boldsymbol W^*;\boldsymbol x_t,\boldsymbol z_t).
    \end{aligned}
\end{equation}

Substituting, we get

\begin{equation}
    \delta \boldsymbol W_{t+1}=(I-\mu H_t)\,\delta \boldsymbol W_t-\mu\,\nabla_{\boldsymbol W}\mathcal L(\boldsymbol W^*; \boldsymbol x_t, \boldsymbol z_t)+o(\|\delta \boldsymbol W\|)\mu.
\end{equation}

We now take the outer product of the recursion and apply the average over $(\boldsymbol x_t,\,\boldsymbol z_t)$. Since $\boldsymbol W^*$ minimizes the population loss, $\mathbb E[\nabla_{\boldsymbol W}\mathcal L(W^*;\boldsymbol x_t,\boldsymbol z_t)]$ vanishes, while $\mathbb E[H_t]=H$ and $\mathbb E[\nabla_{\boldsymbol W}\mathcal L(\boldsymbol W^*;\boldsymbol x_t,\boldsymbol z_t)\nabla_{\boldsymbol W}\mathcal L(\boldsymbol W^*;\boldsymbol x_t,\boldsymbol z_t)^T]=\Sigma_{\text{grad}}$ define the population Hessian and gradient covariance at $\boldsymbol W^*$, and do not depend on $t$. 

We work to leading order in the learning rate: we posit the self-consistent scaling $\mathbb E[\delta \boldsymbol W\delta \boldsymbol W^T]\sim\mathcal O(\mu)$, which is observed empirically and verified a posteriori. This is what allows us to drop the terms $o(\|\delta \boldsymbol W\|)$. The term $\mu^2\,\mathbb E[H_t\,\delta \boldsymbol W_t\delta \boldsymbol W_t^T\, H_t]$ is $\mathcal O(\mu^3)$ and therefore dropped, removing any dependence on Hessian fluctuations at this order. The linear cross-term $\delta \boldsymbol W_t\,\mathbb E[\nabla_{\boldsymbol W}\mathcal L(\boldsymbol W^*)^T]$ vanishes identically, but the term $\mu^2\,\mathbb E[H_t\,\delta \boldsymbol W_t\,\nabla_{\boldsymbol W}\mathcal L(\boldsymbol W^*)^T]$ does not, since $H_t$ and $\nabla_{\boldsymbol W}\mathcal L(\boldsymbol W^*)$ are correlated through a shared sample.

We get a deterministic recursion for the matrix $C_t:=\mathbb E[\delta \boldsymbol W_t\delta \boldsymbol W_t^T]$:

\begin{equation}C_{t+1}=C_t - \mu (HC_t+C_t H^T)+\mu^2\,\Sigma_\text{grad} + O(\mu^2)\,\mathbb E[\delta \boldsymbol W_t] + \mathcal O (\mu^3).
\end{equation}

The term $\mathbb E[\delta \boldsymbol W_t]$ decays as $\mathbb E[\delta \boldsymbol W_{t+1}]=(I-\mu H)\,\mathbb E[\delta \boldsymbol W_t]\to0$ under $\rho(I-\mu H)<1$. We now impose stationarity, namely for $t\to\infty$, $C_{t+1}=C_t:=C$. We get

\begin{equation}HC+CH=\mu\,\Sigma_{\text{grad}},\label{eq:ou}\end{equation}

which is the corresponding Lyapunov equation of our process, where we used $H=H^T$. Notice that $H$ and $\Sigma_{\text{grad}}$ are fixed by the choice of network architecture and teacher weights.

\subsection{Simplifying the Lyapunov equation}\label{app:lyap2}

Computing them at $\boldsymbol W^*$, our equation simplifies further, and we get 

\begin{equation}\begin{aligned}H&=\alpha M_{\text{main}}+\beta M_{\text{aux}}.\\ \Sigma_{\text{grad}}&=\sigma^2\big(\alpha^2 M_{\text{main}}+\beta^2 M_{\text{aux}}\big).\end{aligned}\end{equation}

where $M_{i}:=\mathbb E[\nabla_{\boldsymbol W} y_{i}\nabla_{\boldsymbol W} y_{i}^T]|_{\boldsymbol W^*}$.

Both $H$ and $\Sigma_{\text{grad}}$ are built from the same two matrices. From these, it immediately follows that the Lyapunov equation can be written as

\begin{equation}
\begin{aligned}
\label{eq:gen_lyap}
(\alpha M_{\text{main}}+\beta M_{\text{aux}})\,C+C\,(\alpha M_{\text{main}}+\beta M_{\text{aux}})=\mu\sigma^2(\alpha^2M_{\text{main}}+\beta^2M_{\text{aux}}).
\end{aligned}
\end{equation}

The generalization errors can also be expanded around $\boldsymbol W^*$, obtaining

\begin{equation}
\epsilon_i=\mathbb E[\delta \boldsymbol W^T M_i \delta \boldsymbol W]=\operatorname{tr} (M_i C).
\end{equation}

One can numerically solve Equation \ref{eq:gen_lyap} for $C$ and recover the theoretical generalization error, but $C$ is still a $KN\times KN$ matrix.

Let us focus on $M_\text{main}$. Taking the explicit derivative, it can be written as

\begin{equation}(M_\text{main})_{ik,jl}=\mathbb E\Big[v_iv_j\,g'(h_i)g'(h_j)\,\frac{x_kx_l}{N}\Big].\end{equation}

At large $N$ the input points decorrelates from the activations, $\mathbb E[x_kx_l/N]\to\delta_{kl}/N$, so

\begin{equation}
\begin{aligned}M_\text{main}&=\frac1N\,P_\text{main}\otimes I_N,\\ (P_\text{main})_{ij}&=v_iv_j\,\mathbb E[g'(h_i)g'(h_j)]=v_iv_j\,J_2(ij).
\end{aligned}
\end{equation}

In general, because $M_j=P_j\otimes I_N/N$, the Lyapunov equation inherits the Kronecker structure: $C=\tilde C\otimes I_N$ and it reduces to the $K\times K$ equation

\begin{equation}\begin{aligned}(\alpha P_\text{main}+\beta P_\text{aux})\tilde C+\tilde C(\alpha P_\text{main}+\beta P_\text{aux})=\mu\sigma^2(\alpha^2P_\text{main}+\beta^2P_\text{aux}).\end{aligned}\end{equation}

The generalization error is then

\begin{equation}
    \epsilon_i=\operatorname{tr}(P_i\tilde C).
\end{equation}

\subsection{A consideration on the structure of P}

Now evaluate $J_2$ at the fixed point $W^*$. There, $Q=T=I$ (the student aligns with the teacher), so the $h_i$ are $i.i.d.$ standard normal and

\begin{equation}J_2(ij)=\mathbb E[g'(h_i)g'(h_j)]=\begin{cases}m^2 & i\neq j\\m^2+\kappa & i=j\end{cases}\end{equation}

where $\kappa:=\operatorname{Var}[g'(h)]$ and $m=\mathbb E[g'(h)]$. We can write $J_2=m^2\mathbf 1\mathbf 1^T+\kappa I$, and

\begin{equation}P_\text{main}=m^2\,\boldsymbol v\boldsymbol v^T+\kappa\,D_v\,,\qquad D_v=\operatorname{diag}(v_i^2),\end{equation}

with $P_\text{aux}$ identical under $\boldsymbol v\to\boldsymbol u$. The linear model has $\kappa=0$, from which $P$ is rank-one, and it reduces to a $2\times2$ problem in $\operatorname{span}(\boldsymbol v,\boldsymbol u)$.

\subsection{A discussion of the optimal beta}\label{app:beta_star}

In general, the matrices $P$ do not commute, and there is no mode decomposition, but this analysis shows where $\beta^*$ lies in this simplified setting. If $P_\text{main}$ and $P_\text{aux}$ commute, they are jointly diagonalizable. The equation decouples completely into scalars,

\begin{equation}\begin{aligned}&2(\alpha a_i+\beta b_i)\,\tilde C_{ii}=\mu\sigma^2(\alpha^2 a_i+\beta^2 b_i),\\ &\tilde C_{ii}=\frac{\mu\sigma^2}{2}\frac{\alpha^2 a_i+\beta^2 b_i}{\alpha a_i+\beta b_i},\end{aligned}\end{equation}

where $a_i$ and $b_i$ are the eigenvalues of the two matrices $P$.

With $\alpha=1$ and $\epsilon_{\rm main}=\operatorname{tr}(P_1\tilde C)=\sum_i a_i\tilde C_{ii}$, we get:

\begin{equation}\frac{\epsilon_{\rm main}}{\epsilon_0}=\frac{\sum_i a_i\,\dfrac{a_i+\beta^2 b_i}{a_i+\beta b_i}}{\sum_i a_i}\end{equation}

with $\epsilon_0={\mu\sigma^2}\sum_i a_i/{2}={\mu\sigma^2}\operatorname{tr}P_\text{main}/{2}$, as discussed in Observation 2.

Define per-mode $f_i(\beta)=(a_i+\beta^2b_i)/(a_i+\beta b_i)$ and $r_i=b_i/a_i$. If there is only one mode, then to compute the optimum one can impose $f_i'=0\Rightarrow r_i\beta^2+2\beta-1=0$, so

\begin{equation}\beta^\star(r)=\frac{\sqrt{1+r}-1}{r}\in\big(0,\dfrac12\big),\end{equation}

being $\sqrt2-1$ when $r=1$. The collective optimum solves $\sum_i a_i f_i'(\beta)=0$. We get $\beta^*<1/2$, which aligns with our empirical observation in Figure \ref{fig:improv_theo}.

For the linear model, the two rank-one matrices $P$ generally do not commute, but the problem is exactly solvable, and one can obtain the same results as in Appendix \ref{app:linear}.

\subsection{Effects of higher-order terms}\label{app:high-order}

\begin{figure}[!ht]
    \centering
    \includegraphics[width=0.55\linewidth]{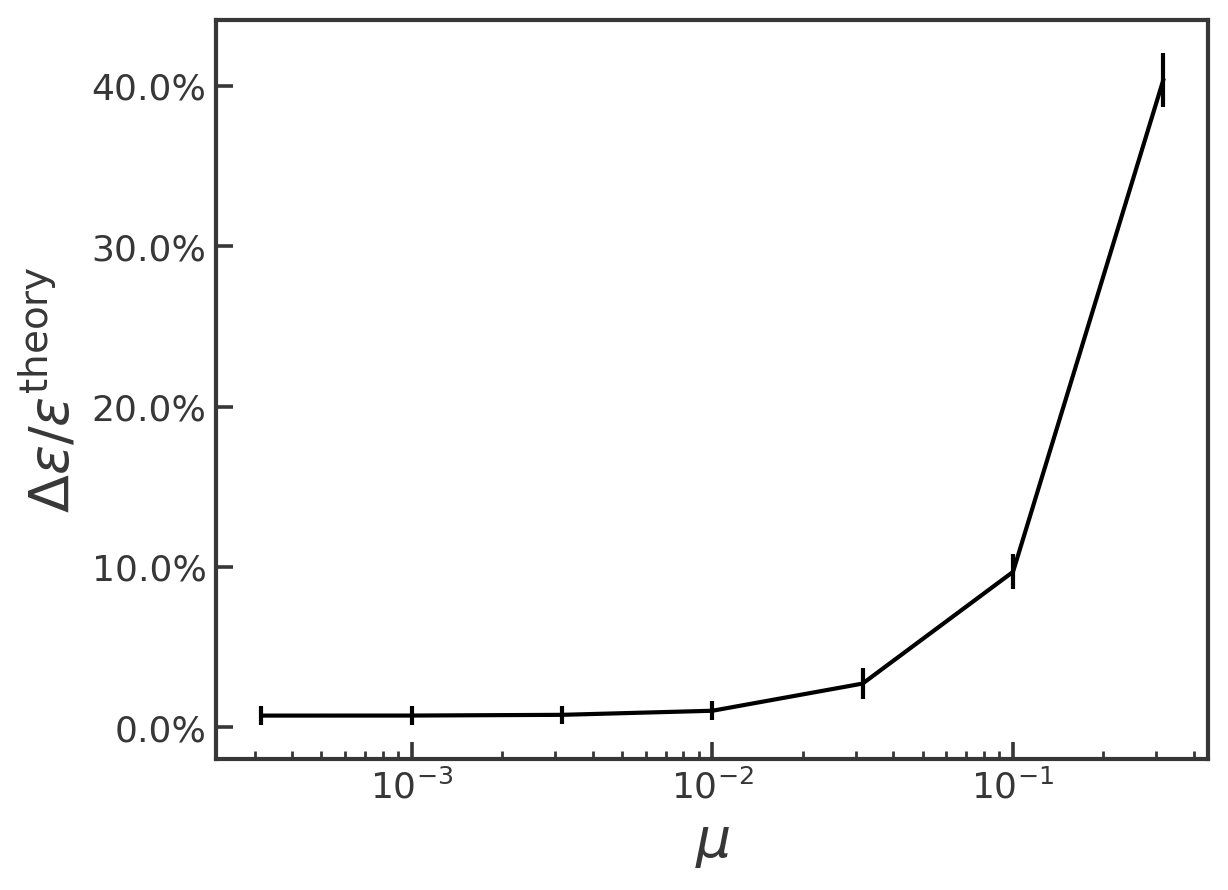}
    \caption{\textbf{Higher-order effects in $\mu$.} Relative error $|\epsilon^\text{ode}-\epsilon^\text{theory}|/\epsilon^\text{theory} = \Delta \epsilon/\epsilon^\text{theory}$ for different $\mu$, with $\epsilon^\text{theory}$ obtained by numerically solving the corresponding Lyapunov equation, and $\epsilon^\text{ode}$ by integrating the ODEs. Average over 10 random choices of $\boldsymbol u$, with erf activation, $\alpha=1$, $\beta=0.5$, $\sigma=0.1$, $K=4$, $\boldsymbol v = [1,1,1,1]$, and correlation between tasks $\rho=0.6$.}
    \label{fig:mu2}
\end{figure}

In Figure \ref{fig:mu2}, we show the impact of higher-order terms on the theory developed above. We choose a set of parameters, namely the activation, $\boldsymbol v$, $\boldsymbol u$, $\beta$, and $\sigma$. We compare the theoretical improvement $\epsilon_{\text{main}}$ predicted by the theory to the one obtained by numerically integrating the ODEs for different choices of $\mu$. As we see, they agree for small $\mu$, while for learning rates approaching $\mu=1$ the higher-order effects modify the results.

\subsection{Convergence of different initializations}\label{app:convergence}

\begin{figure}[!ht]
    \centering
    \includegraphics[width=0.8\linewidth]{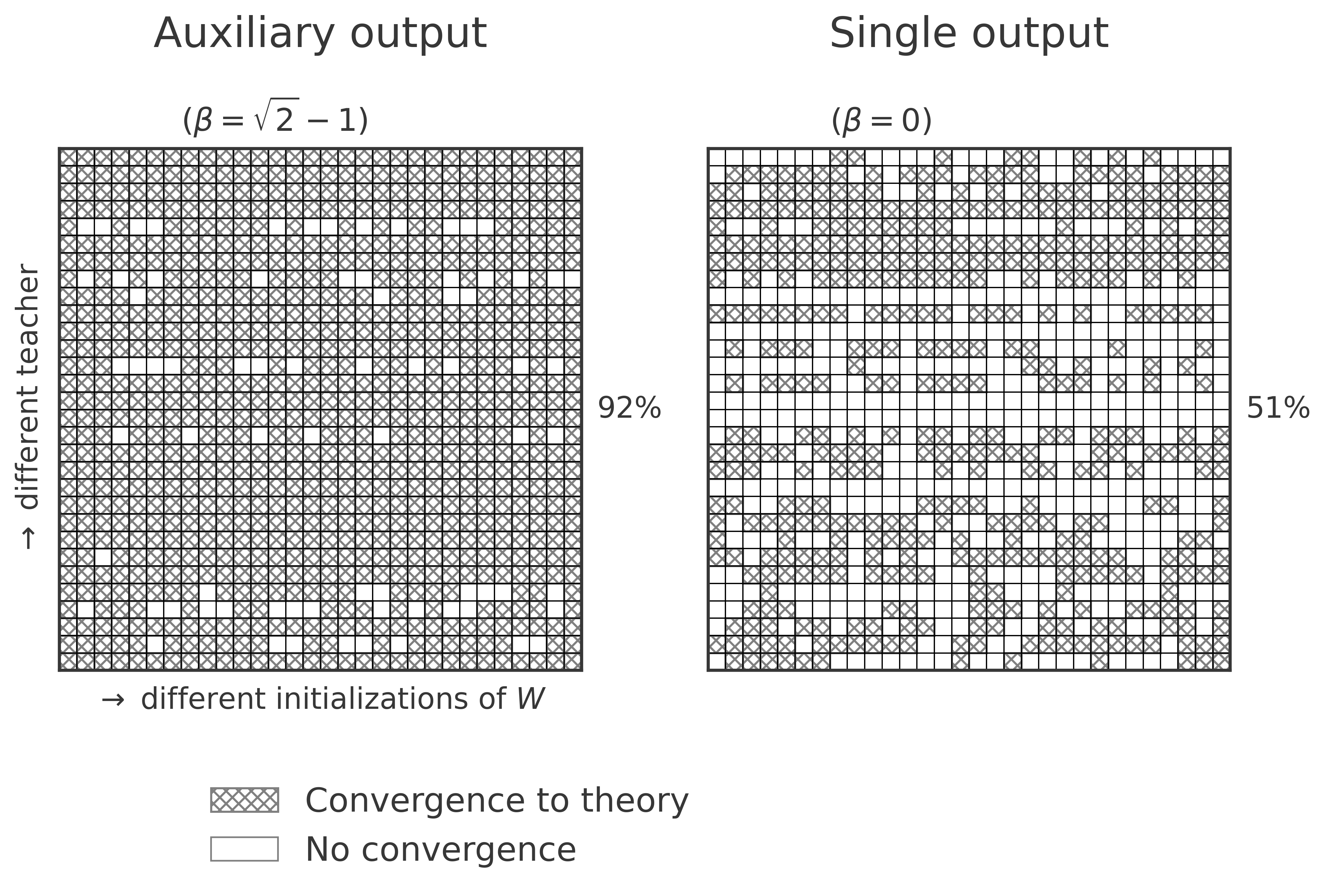}
    \caption{\textbf{Convergence of different initializations.} Comparison of how often ODE integration of different initializations of $W$, $W^*$, $\boldsymbol v$, $\boldsymbol u$ converges to the theoretical $\epsilon^\text{theory}$, when $\beta=\sqrt 2 -1$ and when $\beta=0$. We sample entries independently from a standard normal distribution. Every row is the same teacher initialization of $\boldsymbol v$, $\boldsymbol u$, and $W^*$; every column is the same initialization of $W$. We plot 30 random initializations of $W$ for each of 30 teachers, with erf activation, $\alpha=1$, $\mu=0.01$, $\sigma=0.1$, $K=4$. Convergence when the relative error is less than 5\%.}
    \label{fig:convergence}
\end{figure}

In Section \ref{sec:nonlinear}, we discussed how not all initializations of $W$ converge to a neighbourhood of $W^*$ at the end of the dynamics. In this section, we sketch a preliminary exploration of this phenomenon. We sample $W^*_{ij}, \, W_{ij},\, v_i,\, u_i \sim \mathcal N(0,1)$, and check how often the ODE dynamics error matches the theoretical generalization obtained by numerically solving the corresponding Lyapunov equation.

\begin{table}[!ht]
\centering
\begin{tabular}{ c|c|c c| }
 \multicolumn{2}{c}{} & \multicolumn{2}{c}{$\beta=0$} \\
 \cline{3-4}
 \multicolumn{2}{c|}{} & converges & does not converge \\
 \cline{2-4}
 \multirow{2}{*}{$\beta=\sqrt 2 -1$}
   & converges         & 451 & 374 \\
   & does not converge & 6 & 69 \\
 \cline{2-4}
\end{tabular}
\vspace{0.5em}
\caption{\textbf{Convergence of different initializations.} Paired counts for the runs in Figure \ref{fig:convergence}. McNemar exact test, $p<10^{-10}$.}
\label{tab:convergence}
\end{table}

As seen in Figure \ref{fig:convergence}, more initializations converge to the global minimum $W^*$ when $\beta=\sqrt 2 -1$ than when there is no auxiliary output, namely when $\beta=0$. Table \ref{tab:convergence} shows the counts. Applying McNemar's exact test, we reject the null hypothesis that the marginal convergence probabilities with the two $\beta$ values are the same, with $p<10^{-10}$. This suggests that auxiliary learning can help escape bad minima in the teacher-student setup.

The choice of $\beta=\sqrt2-1$ for the auxiliary output is independent of $\boldsymbol v$, $\boldsymbol u$ and the related optimal $\beta^*$. This analysis could be expanded to more $\beta$ values, but we do not expect major differences.

\section{A different scaling of the main task}\label{app:second-case}

In this section, we study an alternative weighting of the main task, namely setting $\alpha=1-\beta$. Although this may seem similar to choosing $\alpha=1$, because online SGD results are intrinsically dynamical, this choice leads to completely different phenomenology. 

Equation \ref{eq:effective_lr} shows that, for perfectly correlated outputs, the effective learning rate $\mu_{\text{eff}}=\mu(\alpha+\beta)$. From our choice of $\alpha$ then it follows that $\mu_\text{eff}=\mu$. But this choice of $\alpha$ implies that, when the tasks are not perfectly correlated, changing $\beta$ also diminishes the learning rate for the main task. Thus, larger $\beta$ values are beneficial because they reduce the learning rate on the main task. As we will see, $\beta^*$ is always at least $1/2$ with this weighting (and, unlike with $\alpha=1$, $\beta$ has to be smaller than 1). In practice, for the linear case, from Equation \ref{err_linear}, the generalization error on the main task is

\begin{equation}
    \epsilon_{\text{main}}\Big|_{\alpha=1-\beta}=\epsilon_0\bigg( 1-\beta-\rho^2\beta(1-2\beta)\bigg),
\end{equation}

minimized by

\begin{equation}
\label{eq:opt_beta}
    \beta^*=\dfrac{1+\rho^2}{4\rho^2}.
\end{equation}

\begin{figure*}[!ht]
    \centering
    \includegraphics[width=.88\linewidth]{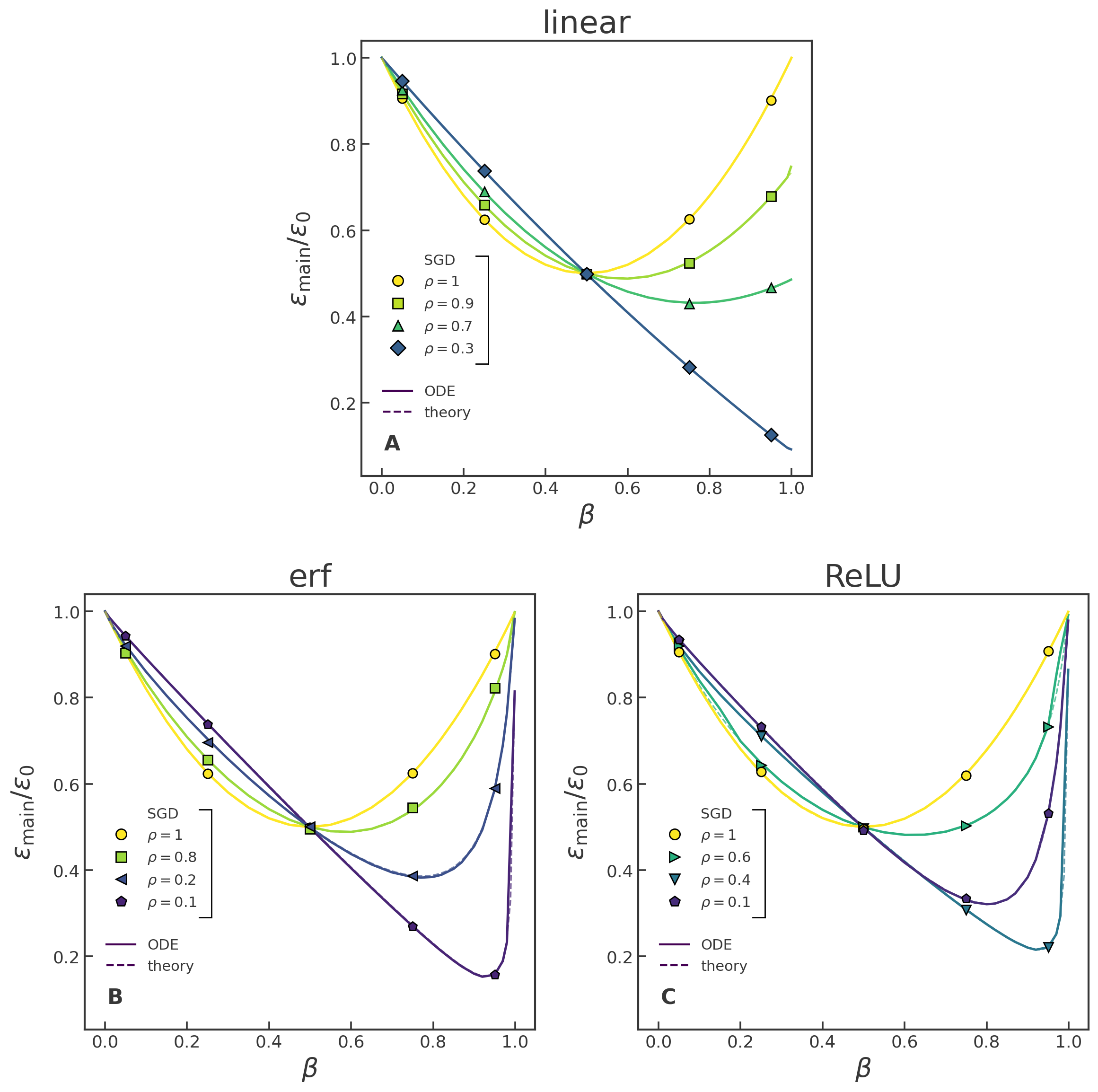}
    \caption{\textbf{Performance of models with $\alpha=1-\beta$.} Relative error $\epsilon_{\text{main}}/\epsilon_0$ for a model with A) linear, B) erf activation, and C) ReLU activation, and different $\rho$ values trained with online gradient descent, $\alpha=1-\beta$, and different $\beta$ values. The dashed lines show the theoretical solutions, the continuous lines show the results of numerically solving the mean-field ODEs, the dots show simulation results for models trained with online SGD, averaged over 10000 epochs after convergence across 5 simulations for linear and erf and 10 for ReLU, with the vertical bars denoting the corresponding confidence intervals. $\mu=0.01$, $\sigma=0.1$, $N=784$, $K=2$ for ReLU and $K=4$ for linear and erf.}
    \label{fig:act_app}
\end{figure*}

Notice that if $\rho=1$, the main and auxiliary tasks are the same, so $\beta^*=1/2$: both tasks should have equal weights in the training loss, but generalization still improves because the noise on each task is independent. When $\rho=0$, then the two tasks are completely decoupled, and rescaling $\alpha$ is effectively just rescaling the learning rate $\mu$. This follows from Equation \ref{eq:pq_dynamics} of Appendix \ref{app:linear} when setting $\rho=0$. Since we must have $\beta < 1$ (otherwise $\alpha$ would be zero or negative), from Equation \ref{eq:opt_beta} we see that for all $\rho<1 / \sqrt 3$ the optimal training setup is the limit $\beta \to 1$. We show results for the linear model in Figure \ref{fig:act_app}A.

When moving from linear to nonlinear models, namely with erf and ReLU activations in Figure \ref{fig:act_app}B--C, we can make some empirical observations. Once again, the behavior does not depend solely on the correlation $\rho$ between $\boldsymbol{v}$ and $\boldsymbol{u}$. We can make a few empirical observations:

\paragraph{Observation 1.} \textit{When $\alpha=\beta=1/2$, then $\epsilon_\text{main}=\epsilon_\text{aux}={\epsilon_0}/2$ for each $\rho$.} 

\paragraph{Observation 2.} $\alpha \epsilon_{\text{main}}+\beta \epsilon_{\text{aux}} = \epsilon_0(\alpha^2+\beta^2)$.

\paragraph{Observation 3.} \textit{The optimal weight $1/2\leq\beta^*<1$.}

These observations match those made with $\alpha=1$ and have a corresponding theoretical explanation.

\end{document}